\documentclass[11pt]{article}
\newtheorem{definition}{Definition}[section]

\newtheorem{example}{Example}[section]
\newtheorem{theorem}{Theorem}[section]
\newtheorem{corollary}{Corollary}[section]

\newcommand{\argmax}{\mathop\textrm{argmax}\limits}
\newcommand{\argmin}{\mathop\textrm{argmin}\limits}

\usepackage{graphicx}
\usepackage{amsmath}
\usepackage{ascmac}
\usepackage{amssymb}
\usepackage{algorithmic}
\usepackage{algorithm}
\usepackage{bm}
\usepackage{color}
\usepackage{float}
\usepackage{tabularx}
\usepackage{comment}
\usepackage{here}
\title{Descriptive Dimensionality and Its Characterization of\\
 MDL-based Learning and Change Detection}

\author{
        Kenji Yamanishi
\thanks{%
K. Yamanishi belongs to
Graduate School of Information Science and Technology, The University of Tokyo,
7-3-1 Hongo, Bunkyo-ku, Tokyo, JAPAN. 
E-mail: yamanishi@mist.i.u-tokyo.ac.jp \protect\\
}}

\begin{document}
\setlength{\abovedisplayskip}{2pt}
\setlength{\belowdisplayskip}{2pt}

\maketitle

\begin{abstract}
This paper  introduces a new notion of dimensionality of probabilistic models from an information-theoretic view point.
We call it the {\em descriptive dimension}~(Ddim).
We show that Ddim coincides with the number of independent  parameters for the parametric class, and can further be extended to real-valued dimensionality when a number of models are mixed.
The paper then derives the rate of convergence of the MDL (Minimum Description Length) learning algorithm which outputs a normalized maximum likelihood (NML) distribution with model of the shortest NML codelength.
The paper proves that the rate is governed by Ddim.
The paper also derives error probabilities of the  MDL-based test for multiple model change detection. It proves that they  are also governed by Ddim.
Through the analysis, we demonstrate that Ddim is an intrinsic quantity which characterizes the  performance of the MDL-based learning and change detection.

\end{abstract}

\section{Introduction}

\subsection{Motivation}

We are concerned with model dimensionality of a probabilistic model. Model dimensionality usually means the number of independent parameters, which we call the {\em parametric dimensionality}. For example, it is the number of real-valued parameters in the regression model, the number of mixture components in the finite mixture model, etc.
This paper reinterprets the model dimensionality from an information-theoretic viewpoint, specifically on the basis of the theory of the minimum description length (MDL) principle~\cite{rissanen78}.

Our motivation is summarized as follows:
First, model dimensionality is an important notion to understand how complex a given model class is. We may measure this complexity in terms of description length from an information-theoretic viewpoint.
This can be calculated no matter whether the model class is parametric or non-parametric, or no matter whether a number of model classes are synthesized or not.
Then  model dimensionality would not necessarily be integer-valued.
We are thus interested in the problems of how we can formalize the description-based model dimensionality in a general way and how it is related to the conventional parametric dimensionality.

Secondly, the complexity of a model class is closely related to the performance of algorithms learning it.
Specifically,  the MDL-based learning algorithm~\cite{barron, yamanishi92, rissanen, kawakita} and the MDL-based change detection algorithms~\cite{ym07, yamanishi2, yamanishi19}, which have been designed on the basis of the MDL principle,  have turned out to be effective in the scenario of machine learning and information theory.
We can infer that their rates of convergence or exponents in error probabilities  would possibly be characterized in terms of description-based model dimensionality.
This is analogous to the fact that the rates of convergence of the empirical risk minimization algorithms are characterized by Vapnik-Chervonenkis dimension~\cite{blumer,vapnik} or metric dimension of a function class~\cite{pollard,Haussler}.
We are thus interested in analyzing the relations among the description-based dimensionality and the performance of the MDL-based learning and change detection algorithms.

The purpose of  this paper is twofold:
One is to introduce a new notion of description-based model dimensionality.
We call this notion the {\em descriptive dimensionality}~(Ddim).
The classical integer-valued parametric dimensionality is extended into a real-valued one for the case where a number of model classes of different complexities are mixed.
The other purpose is to derive new theoretical results on the MDL-based algorithms for learning and change detection.
We thereby characterize them in terms of Ddim.

\subsection{Related Work}

A number of notions of dimensionality have been proposed
in the areas of physics and statistics.
The metric dimension was proposed by Kolmogorov and Tihomirov~\cite{kol} to measure the complexity of a given set of points on the basis of the notion of covering numbers. This was evolved to the notion of  the box counting dimension, equivalently, the fractal dimension~\cite{man, farmer}.
It is also related to the capacity~\cite{dudley}.
The metric dimension was used to measure the complexity of a class of functions and was related to the rate of uniform convergence over the class~(see e.g. \cite{dudley}, \cite{pollard}).
Vapnik Chervonenkis (VC) dimension was proposed to measure the power of representation for a given class of functions~\cite{vapnik}. It was also related to the rate of uniform convergence of the algorithm for minimizing empirical losses over the class.
See \cite{Haussler} for relations between dimensionality and learning.
In all of the previous work on dimensionality,  it has not been related to information-theoretic notions such as codelengths.

As for the MDL principle, Rissanen has proven the consistency of the model estimated by the MDL criterion~\cite{rissanen89}.
In earlier stages of MDL research, the {\em two-stage MDL} distribution has extensively been studied, that is, the data and model are encoded in two steps, then the probability distribution with model of the shortest two-stage codelengths have been analyzed.
Recently, the normalized maximum likelihood (NML) distribution has received more attentions than the two-stage MDL distribution. This is because the NML distribution achieves the minimum of Shtarkov's minimax regret (see Section 2.1), and
the NML distribution with model of the shortest codelength has estimation optimality~\cite{rissanen}.  See \cite{grunwald}  for recent advances of MDL.

The MDL principle has extensively been applied to learning theory and change detection.
As for learning, i.e, model estimation,
Barron and Cover first showed the rate of convergence of the MDL model estimator and characterized it in terms of index resolvability~\cite{barron}.
Note that they evaluated  the model with the shortest two-stage codelength over the discretized model class.
Yamanishi gave a non-asymptotic form of the rate of convergence of the model with the two-stage shortest codelength in the PAC~(probably approximately correct) learning framework~\cite{yamanishi92}.
He focused on the specific model called the stochastic rules with finite partitioning, which can be thought of as piecewise Bernoulli models. In \cite{barron, yamanishi92}, the model class was discretized and the convergence was analyzed over the discretized class.
Chatterjee and Barron~\cite{cb} and Kawakita and Takeuchi~\cite{kawakita} designed the MDL learning algorithms without discretization and applied them to the problem of supervised learning with L1-regularization. In their analysis,  the target of analysis remained to be models of the shortest two-stage codelength, and the discretization technique still played an essential role in their analysis.

As for model change detection, Yamanishi and Maruyama formulated the problem of {\em dynamic model selection}~(DMS) when the underlying probabilistic model changes over time~\cite{ym05, ym07}.  DMS is to detect time-varying models from a stream data. They developed the MDL-based DMS algorithm to solve the problem.
The issues similar to DMS have been addressed in a number of different scenarios including switching
distributions~\cite{erven}, derandomization~\cite{vovk}, tracking best experts~\cite{herbster}, Bayesian change point detection in multivariate time series~\cite{xuan}, concept drift~\cite{gama}, structural break estimation for autoregression models~\cite{davis}. etc.
Vreeken et al. proposed Krimp as a description length-based test statistics for change point detection and demonstrated its empirical effectiveness~\cite{vreeken}.
Yamanishi and Fukushima~\cite{yamanishi19} developed the MDL-based model change statistics using the NML codelength to propose a hypothesis testing for model change detection, which we call the MDL test.
They derived non-asymptotic forms of Type I and II error probabilities of the MDL  test, both of which converge to zero exponentially in data length.

\subsection{Significance of This Paper}

The significance of this paper is summarized as follows:\\
(1){\em Model dimensionality is reformulated through Ddim.}
We define Ddim similarly with the box counting dimension in that it is given by the logarithm of $\epsilon$-covering number divided by $\log (1/\epsilon )$.
 Hence Ddim can be defined regardless of whether the model class is parametric or non-parametric, as with the box counting dimension.
However, Ddim is unique in that the covering number for Ddim is defined as the least number of representative points necessary for approximating the shortest codelength for the model class.
 We show that Ddim coincides with the parametric dimensionality when the model class is a single parametric class. Ddim can be further defined for the cases where a number of parametric model classes are probabilistically mixed (model fusion) or are concatenated (model concatenation). Then Ddim can be real-valued.
Such cases may occur e.g. when the model changes over time.

2){\em Theoretical results on the performance of MDL-based learning and change detection are updated to be characterized by Ddim.}
We derive the rate of convergence of MDL learning algorithm both for the cases where the true distribution is in a given family of model classes and is not in the family.
 In previous work on MDL learning~\cite{barron, yamanishi92,cb,kawakita}, the maximum likelihood  distribution with model with the shortest two-stage codelength were analyzed.
Meanwhile, this paper derives upper bounds on the rate of convergence of the NML distribution with model of the shortest NML codelength.
Unlike all of the existing work, we do not use any discretization technique both in the design and analysis of MDL learning.
The result gives a new justification of MDL learning and gives a new rationale for the NML distribution.
To the author's knowledge, this is the first result on the rate of convergence on the NML distribution.
We relate the rate of convergence of the NML distribution to Ddim for the true model.
Furthermore, in the case where a number of models are probabilistically mixed, we prove that the rate of convergence of the NML distribution is governed by Ddim for model fusion.

We also consider the problem of model change detection.
We conduct a hypothesis testing whether a data sequence comes from a a given model sequence or not.
The model sequence may include multiple change points.
This is a scenario different from \cite{yamanishi19}, in which it has been tested whether a model change has occurred or not at a given single change point.
We propose the MDL change statistics for this new scenario, on the basis of the notion of dynamic model selection~\cite{ym05}.
We
derive upper bounds on Type I and II error probabilities of hypothesis testing for the MDL-based model change detection.
We prove that the error probabilities decay exponentially to zero as sample size increases, and that their exponents are governed by Ddim for model concatenation.

Through the analysis, we demonstrate that Ddim is an intrinsic quantity which characterizes the  performance of  MDL-based learning and change detection.

The rest of this paper is organized as follows:
Section 2 gives a formal definition of Ddim and its theoretical properties.
Section 3 evaluates rates of convergence of learning the NML distributions and relates them to Ddim.
Section 4 evaluates error probabilities for the MDL-based change detection and relates them to Ddim.
Section 5 gives concluding remarks.
Appendix gives proofs of a number of theorems.

\section{Descriptive Dimensionality}

\subsection{NML and Parametric Complexity}

This section gives a formal definition of Ddim from an information-theoretic viewpoint.
Let ${\mathcal X}$ be the data domain where ${\mathcal X}$ is either discrete or continuous. Without loss of generality, we assume that ${\mathcal X}$ is discrete.
Let
${\bm x}=x_{1},\dots ,x_{n}\in {\mathcal X}^{n}$ be a data sequence of length $n$.
We drop $n$ when it is clear from the context.
We assume that each $x_{i}$ is independently generated.
${\mathcal P}=\{p({\bm x}) \}$ be a class of probabilistic models where $p({\bm x})$ is a probability mass function or a probability density function.
${\mathcal P}$ can be a real-valued or discrete-valued parametrized class. In either case, through the paper, we assume that there exists $\hat{p}=\argmax_{p\in {\mathcal P}}p({\bm x})$ for any ${\bm x}$.
We start by defining the NML codelength, the fundamental notion in the MDL principle.

\begin{definition}{\rm
We define the {\em normalized maximum likelihood (NML) distribution} over ${\mathcal X}^{n}$ relative to ${\mathcal P}$, which we denote as $p_{_{\rm NML}}({\bm x};{\mathcal P})$,  by\\
\begin{eqnarray}\label{nmld}
p_{_{\rm NML}}({\bm x};{\mathcal P})\buildrel \rm def \over =\frac{\max _{p\in {\mathcal P}}p({\bm x})}{\sum _{{\bm y}} \max _{p\in {\mathcal P}}p({\bm y})}. \\
\nonumber
\end{eqnarray}

The {\em normalized maximum likelihood (NML) codelength} of ${\bm x}$ relative to ${\mathcal P}$, which we denote as $L_{_{\rm NML}}({\bm x}; {\mathcal P})$, is given as follows:
\begin{eqnarray}\label{sc0}
L_{_{\rm NML}}({\bm x};{\mathcal P})&\buildrel \rm def \over =&-\log p_{_{\rm NML}}({\bm x}; {\mathcal P})
\nonumber \\
           &=&-\log \max _{p\in {\mathcal P}}p({\bm x})+\log
           {\mathcal C}_{n}({\mathcal P}),
\end{eqnarray}
where
\begin{eqnarray}\label{scc}
\log {\mathcal C}_{n}({\mathcal P})\buildrel \rm def \over =\log
\sum_{{\bm y}} \max _{p\in {\mathcal P}}p({\bm y}).
\end{eqnarray}
}
\end{definition}
The first term in (\ref{sc0}) is the negative logarithm of  the  maximum likelihood while the second term (\ref{scc}) is the logarithm for the normalization term.
The latter is called
the {\em parametric complexity} of ${\mathcal P}$~\cite{rissanen}.
This means the information-theoretic complexity for the model class ${\mathcal P}$.
The NML codelength can be thought of as an extension of Shannon information $-\log p({\bm x})$ into the case where the true model $p$ is unknown but only ${\mathcal P}$ is known.

In order to understand the meaning of the NML codelength and the parametric complexity,  according to \cite{shtarkov},
we define the {\em minimax regret} as follows:\\
\begin{eqnarray*}\label{minimaxregret}
R_{n}({\mathcal P})\buildrel \rm def \over =\min _{q} \max_{{\bm x}}\left\{ -\log q({\bm x})-\min _{p\in {\mathcal P}}(-\log p({\bm x}))\right\},
\end{eqnarray*}\\
where the minimum  is taken over the set of all probability distributions.
The minimax regret means the descriptive complexity of the model class, indicating how largely any prefix codelength is
deviated from the smallest negative
log-likelihood over the model class.
Shtarkov proved that the NML distribution (\ref{nmld}) is optimal in the sense that it attains the minimum of  the minimax regret \cite{shtarkov}. In this sense the NML codelength is the optimal codelength required for encoding ${\bm x}$ for given ${\mathcal P}$.
Then the minimax regret coincides with the parametric complexity~\cite{shtarkov}. That is,\\
\begin{eqnarray}\label{pshtarkov}
R_{n}({\mathcal P})=C_{n}({\mathcal P}).\\ \nonumber
\end{eqnarray}

We next consider how to calculate the parametric complexity $C_{n}({\mathcal P})$.
According to \cite{rissanen} (pp:43-44), the parametric complexity can be represented using a variable transformation technique as follows:
\begin{eqnarray}\label{int1}
 C_{n}({\mathcal P})=
 \sum _{{\bm y}}  \max _{p\in {\mathcal P}}p({\bm y})
         =\int g(\hat{p}, \hat{p})d\hat{p},
\end{eqnarray}
where $g(\hat{p},p)$ is defined as
\begin{eqnarray}\label{gfunc}
g(\hat{p}, p )\buildrel \rm def \over =
\sum
 _{{\bm y}:\max _{\bar{p}\in {\mathcal P}}\bar{p}({\bm y})=\hat{p}({\bm y})}
 p({\bm y}).\\
 \nonumber
\end{eqnarray}
Note that for fixed $p$, $g(\hat{p},p)$ forms a probability density function of $\hat{p}$. That is, \\
\begin{eqnarray*}
\int g(\hat{p} ,p )d\hat{p}=1.
\end{eqnarray*}

\subsection{Definition of Descriptive Dimensionality}
Below we give the definition of Ddim from a view of approximation of the parametric complexity, equivalently, the minimax regret~(see (\ref{pshtarkov})).
The overall scenario of defining Ddim is as follows:
We first count how many points are required to approximate the parametric complexity (\ref{int1}) with discretization of ${\mathcal P}$.
We consider the counts as information-theoretic richness of representation for a  model class.
We then employ that counts to define Ddim in a similar manner with the box counting dimension~\cite{dudley}.

We consider to approximate (\ref{int1}) with a finite sum of partial integrals of $g(\hat{p},\hat{p})$.
Let
$\overline{{\mathcal P}}=\{p _{1}, p _{2},\dots\}
$ 
be a finite subset of 
${\mathcal P}$.
For $\epsilon >0, $ for $p_{i}\in \overline{{\mathcal P}}$, let
\begin{eqnarray*}\\
D_{\epsilon}^{n}(i)\buildrel \rm def \over =\{p\in {\mathcal P} :\ d_{n}(p_{i},p)\leq \epsilon ^{2}\},
\end{eqnarray*}\\ where
$d_{n}(p_{i}, p)$ is the Kullback-Leibler (KL) divergence between $p_{i}$ and $p$: \\
\begin{eqnarray}\label{kl}
d_{n}(p_{i}, p)\buildrel \rm def \over =\frac{1}{n}\sum _{{\bm x}}
p_{i}({\bm x})\log \frac{p_{i}({\bm x})}{p({\bm x})}.
\end{eqnarray}\\
 Then we approximate ${\mathcal C}_{n}({\mathcal P})$ as
\begin{eqnarray}\label{approx}
\overline{{C}_{n}}(\overline{{\mathcal P}})\buildrel \rm def \over = \sum 
_{i}Q_{\epsilon}(i),
\end{eqnarray}
where
\begin{eqnarray*}\label{repp}
Q_{\epsilon}(i)\buildrel \rm def \over =\int _{\hat{p}\in D_{\epsilon}^{n}(i)}g(\hat{p}, \hat{p})d\hat{p}.
\end{eqnarray*}\\
That is, (\ref{approx}) gives an approximation to $C_{n}({\mathcal P})$ with a finite sum of integrals of $g(\hat{p}, \hat{p})$ over the
$\epsilon ^{2}-$neighborhood of a  point $p_{i}$ with respect to the KL-divergence.
We define  $m_{n}(\epsilon :{\mathcal P})$ as the smallest number of  points $|\overline{{\mathcal P}}|$ with respect to $\overline{\mathcal P}$ such that
$C_{n}({\mathcal P}) \leq \overline{C}_{n}(\overline{{\mathcal P}})$. More precisely,\\
\begin{eqnarray}\label{countsdef}
m_{n}(\epsilon :{\mathcal P})\buildrel \rm def \over =\min _{\overline{{\mathcal P}}}
|\overline{{\mathcal P}}|\ \ {\rm subject\ to}\
C_{n}({\mathcal P})\leq \overline{C_{n}}(\overline{{\mathcal P}}).
\end{eqnarray}

We are ready to introduce the descriptive dimension as follows:
\begin{definition} {\rm
Let ${\mathcal P}$ be a class of probabilistic models.
We let $m(\epsilon :{\mathcal P})$ be the one obtained by
 choosing $\epsilon ^{2}n=\Theta (1)$ in $m_{n}(\epsilon :{\mathcal P} )$ as in (\ref{countsdef}).
If the following limit (\ref{defdim}) exists, we define the {\em descriptive dimension}~(Ddim) of ${\mathcal P}$ as\\
\begin{eqnarray}\label{defdim}
{\rm Ddim}({\mathcal P})\buildrel \rm def \over =\lim _{\epsilon \rightarrow 0}\frac{\log m(\epsilon : {\mathcal P})}{\log (1/\epsilon )}. \\
\nonumber
\end{eqnarray}
}
\end{definition}

The definition of Ddim is similar with that of the {\em box counting dimension}~\cite{dudley,man,farmer}.
The main difference between them is how to count the number of points.
 Ddim is calculated on the basis of the number of points required for approximating the parametric complexity,
while the box counting dimension is calculated on the basis of the number of points required for covering a given object with their $\epsilon $-neighborhoods with some metric.

We denote $m_{n}({\mathcal P})$ as the total number of  representative points for parametric complexity for ${\mathcal P}$ obtained by choosing $\epsilon ^{2}n=\Theta (1)$ in $m(\epsilon :{\mathcal P})$ as in (\ref{countsdef}).
Eq.(\ref{defdim}) is then equivalent with\\
\begin{eqnarray}\label{defdim2}
{\rm Ddim}({\mathcal P})=\lim _{n \rightarrow \infty}\frac{2\log m_{n}({\mathcal P})}{\log  n}.
\end{eqnarray}\\
\ \ \ \ In order to verify that Ddim is a reasonable definition of dimensionality, we show that Ddim coincides with the parametric dimensionality in the special case where the model class is a finite dimensional parametric one.

Consider the case where ${\mathcal P}_{k}$ is a $k$-dimensional parametric class, i.e.,
${\mathcal P}_{k}=\{p({\bm x};\theta ):\ \theta \in \Theta _{k}\subset {\mathbb R}^{k}\}$,
where $\Theta _{k}$ is a $k$-dimensional real-valued parameter space.
Let $p({\bm x};\theta )=f({\bm x}|\hat{\theta}({\bm x}))g(\hat{\theta}({\bm x});\theta )$ for the conditional probabilistic mass function $f({\bm x}|\hat{\theta}({\bm x}))$.
We then write  $g$ according to (\ref{gfunc})  as follows \\
\begin{eqnarray*}
g(\hat{\theta}, \theta )=\sum _{{\bm x}:\argmax_{\theta}p({\bm x};\theta)=\hat{\theta}}p({\bm x};\theta).\\
\end{eqnarray*}

Assume that the central limit theorem holds for the maximum likelihood estimator of a parameter vector $\theta$.
Then according to \cite{rissanen}, we can
 take a Gaussian density function as (\ref{gfunc}) asymptotically.
That is, for sufficiently large $n$, (\ref{gfunc}) converges to the Gaussian distribution in law:
\begin{eqnarray}\label{clt}
g(\hat{\theta}, \theta )\longrightarrow \left(\frac{n}{2\pi }\right)^{\frac{k}{2}}|I_{n}(\theta)|^{\frac{1}{2}}e^{-n(\hat{\theta}-\theta)^{\top}I_{n}(\theta )(\hat{\theta}-\theta )/2},\\
\nonumber
\end{eqnarray}
where $I_{n}(\theta )\buildrel \rm def \over =(1/n)E_{\theta}[-\partial ^{2}\log p({\bm x};\theta )/\partial \theta \partial \theta ^{\top}]$ is the Fisher information matrix.

Under the assumption of (\ref{clt}),
the following theorem shows the basic property of $m_{n}(\epsilon :{\mathcal P}_{k})$ for the parametric case.
\begin{theorem}\label{basic}
Suppose that $p({\bm x};\theta )\in {\mathcal P}$ is continuously three-times differentiable with respect to $\theta$.
Under the assumption of the central limit theorem so that (\ref{clt}) holds,
for sufficiently large $n$, we have
\begin{eqnarray}\label{nmb1}
\log C_{n}({\mathcal P}_{k}) = \log m_{n}(\epsilon :{\mathcal P}_{k})+\frac{k}{2}\log (\epsilon ^{2}n)+O(1). \\ \nonumber
\end{eqnarray}
\end{theorem}
(Since the length proof is long for the introduction of Ddim, we move the proof of Theorem \ref{basic} to Appendix A.1.)

Theorem \ref{th1} relates Ddim to the parametric  dimensionality for the parametric case.
\begin{theorem}\label{th1}
For a $k$-dimensional parametric class ${\mathcal P}_{k}=\{p({\bm x};\theta):\ \theta \in \Theta _{k}\subset {\mathbb R}^{k}\}$,  under the assumption as in Theorem \ref{basic} for ${\mathcal P}_{k}$, we have\\
\begin{eqnarray}\label{pcase}
{\rm Ddim}({\mathcal P}_{k})=k. \\
\nonumber
\end{eqnarray}
\end{theorem}
{\em Proof:}
We denote $m_{n}({\mathcal P})$ as that obtained by choosing $\epsilon ^{2}n=\Theta (1)$ in $m(\epsilon :{\mathcal P})$.
According to \cite{rissanen} (p.53),
when the class ${\mathcal P}$ is a $k$-dimensional parametric class ${\mathcal P}_{k}$, then under the central limit theorem condition for the maximum likelihood estimator for each $\theta$, the parametric complexity is asymptotically
expanded as follows:\\
\begin{eqnarray}\label{asymp}
\log C_{n}({\mathcal P}_{k})=\frac{k}{2}\log \frac{n}{2\pi}+\log \int \sqrt{|I(\theta )|}d\theta +o(1),
\end{eqnarray}\\
where $I(\theta )$ is the Fisher information matrix:
$I(\theta )\buildrel \rm def \over =\lim _{n\rightarrow \infty}(1/n){\rm E}_{\theta}[-\partial ^{2}\log p(X^{n};\theta )/\partial \theta \partial \theta ^{\top}]$.
Plugging (\ref{nmb1}) with (\ref{asymp}) into (\ref{defdim2})
yields (\ref{pcase}). This completes the proof of Theorem \ref{th1}.
\hspace*{\fill}$\Box$\\

Theorem \ref{basic} can be generalized no matter whether  the model class is either parametric or non-parametric.
Eq.(\ref{gfunc2}) in the proof of Theorem \ref{basic}~(see Appendix A.1), which relates the central limit theorem to the volume of the discretized mass, is the key to prove Theorem \ref{basic}. By generalizing (\ref{gfunc2}), we have the following form of a relation between $C_{n}({\mathcal P})$ and $m_{n}({\mathcal P})$.

\begin{theorem}\label{basic2}
Suppose that for $\epsilon^{2}n=\Theta (1)$, it holds:\\
\begin{eqnarray}\label{cc2}
\sup _{\overline{\mathcal P}}\sup _{p_{i}\in {\overline{\mathcal P}}}g(p_{i},p_{i})|D_{\epsilon}(i)|=O(1).\\ \nonumber
\end{eqnarray}
Then we have
\begin{eqnarray*}
\log C_{n}({\mathcal P})=\log m_{n}({\mathcal P})+O(1).\\
\end{eqnarray*}
\end{theorem}

Note that (\ref{cc2}) can be thought of as a kind of generalized variants of the central limit theorem.

Theorem \ref{th1} shows that Ddim coincides with the parametric  dimensionality when the model class is a single parametric one.
Ddim can also be defined even for the case where a number of parametric classes are fused or concatenated, as shown below.

\begin{figure}[!thb]
\begin{center}
\begin{tabular}{c}
\begin{minipage}{0.45\hsize}
\begin{center}
\includegraphics[height=28mm
]{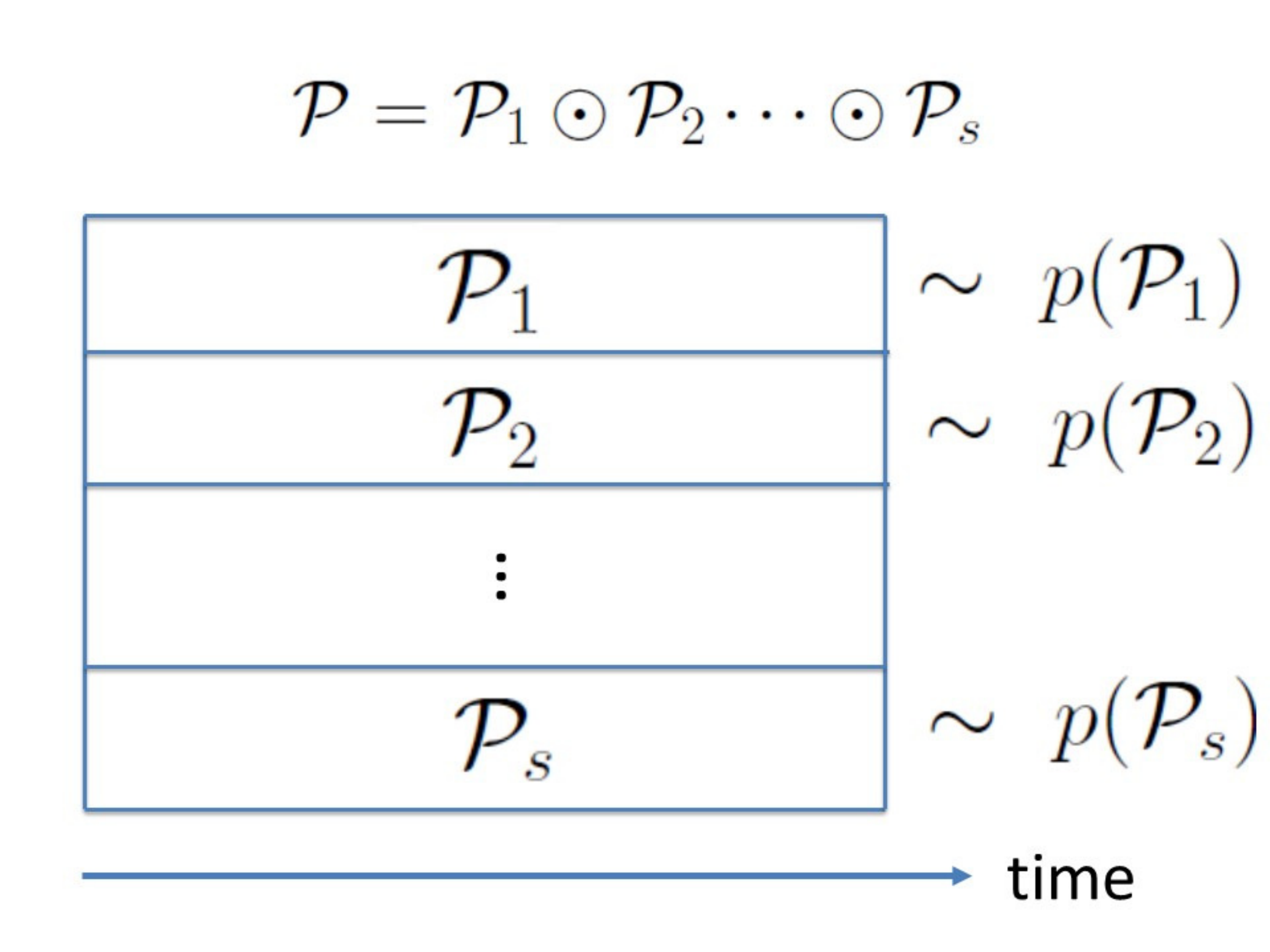}
		\\
(a) Model fusion
\end{center}
\end{minipage}

\begin{minipage}{0.45\hsize}
\begin{center}
\includegraphics[height=28mm
]{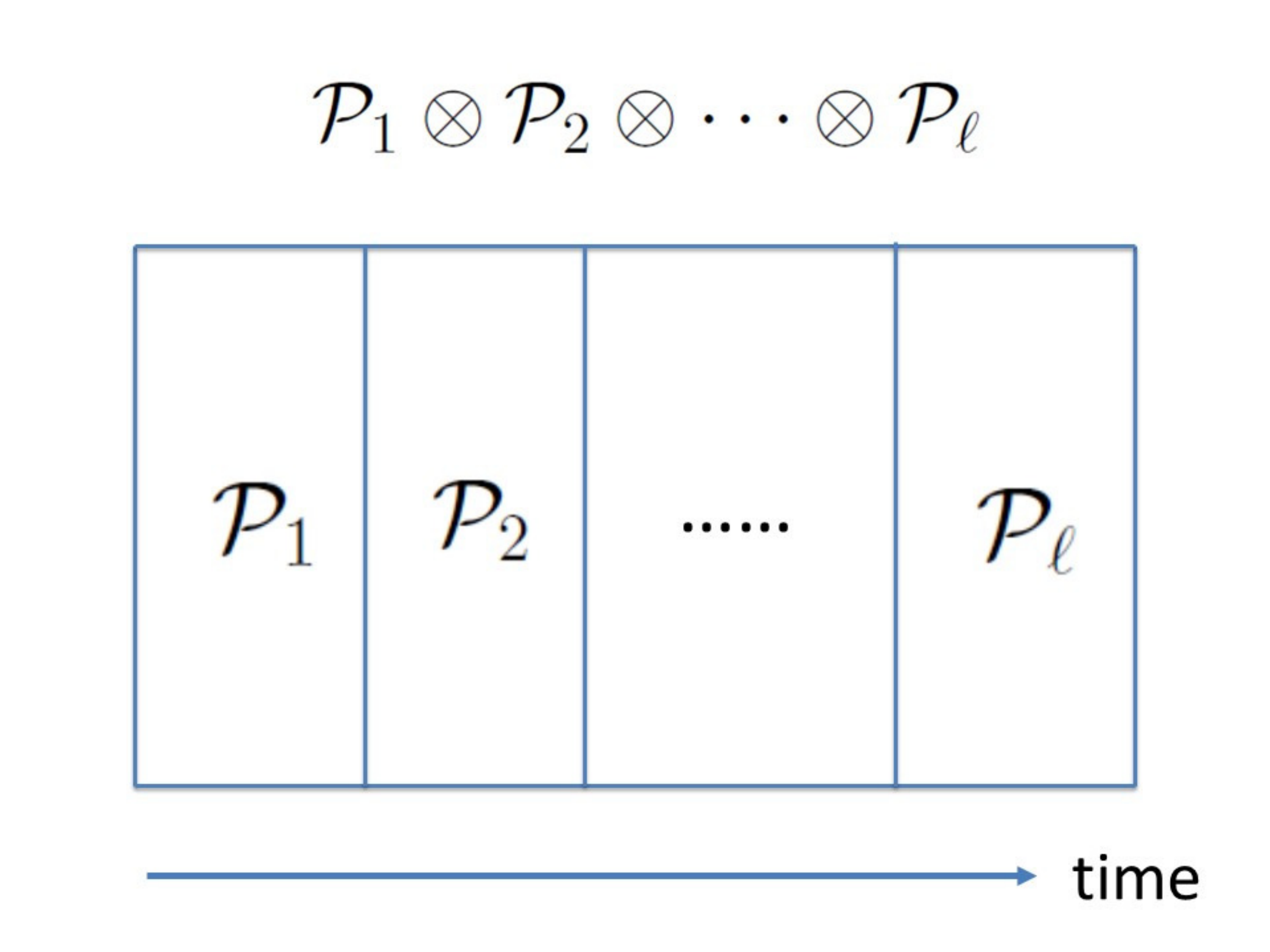}
\\
(b) Model concatenation
\end{center}
\end{minipage}
\end{tabular}
\end{center}
\caption{Model fusion and concatenation}
\label{fig1}
\end{figure}

We first consider {\em model fusion} where a number of model classes are probabilistically mixed, as in Fig.~\ref{fig1} (a).
Let ${\mathcal F}=\{ {\mathcal P}_{1},\dots , {\mathcal P}_{s}\}$ be a family of model classes and assume a model class is probabilistically distributed according to
 $p({\mathcal P})$ over ${\mathcal F}$ where $p({\mathcal P})$ does not change over time.
 The true distribution is determined once ${\mathcal P}$ is generated.
We denote model fusion over ${\mathcal F}$ as ${\mathcal F}^{\odot }={\mathcal P}_{1}\odot \cdots \odot {\mathcal P}_{s}$.

Then by taking the expectation of $m(\epsilon : {\mathcal P})$ with respect to ${\mathcal P}$, the definition of Ddim of ${\mathcal F}^{\odot}$ is naturally induced as follows:\\
 \begin{eqnarray}\label{fusiondef}
{\rm Ddim} ({\mathcal F}^{\odot})&\buildrel \rm def \over =&\lim _{\epsilon \rightarrow 0}\frac{\log E_{{\mathcal P}}[m(\epsilon : {\mathcal P})]}{\log (1/\epsilon )}.\\
\nonumber
 \end{eqnarray}

We immediately obtain the following lower bound on Ddim of model fusion.
\begin{theorem}\label{fusionddim}
\begin{eqnarray}\label{fusionlower}
{\rm Ddim} ({\mathcal F}^{\odot})\geq \sum_{i=1}^{s}p({\mathcal P}_{i}){\rm Ddim}({\mathcal P}_{i}).
\end{eqnarray}
\end{theorem}
{\em Proof.} We employ the Jensen's inequality to obtain
\begin{eqnarray*}
{\rm Ddim} ({\mathcal F}^{\odot})&=&\lim _{\epsilon \rightarrow 0}\frac{\log E_{{\mathcal P}}[m(\epsilon : {\mathcal P})]}{\log (1/\epsilon )}\\
&\geq &\lim _{\epsilon \rightarrow 0}\frac{E_{{\mathcal P}}[\log m(\epsilon : {\mathcal P})]}{\log (1/\epsilon )}\\
&=&\sum_{i=1}^{s}p({\mathcal P}_{i})\lim _{\epsilon \rightarrow 0}\frac{\log m(\epsilon :{\mathcal P}_{i})}{\log (1/\epsilon )}\\
&=&\sum_{i=1}^{s}p({\mathcal P}_{i}) {\rm Ddim}({\mathcal P}_{i}).
\end{eqnarray*}
This completes the proof of Theorem \ref{fusionddim}.
\hspace*{\fill}$\Box$\\

\begin{definition}{\rm
We define the {\em pseudo Ddim}; $\underline{{\rm Ddim}}({\mathcal F}^{\odot})$ for model fusion over ${\mathcal F}$ as the righthand side of  (\ref{fusionlower}).}
\end{definition}

\begin{example}{\rm
Let $p({\mathcal P})$ be a prior distribution of ${\mathcal P}$ over ${\mathcal F}$.
When a data sequence ${\bm x}=x_{1},\dots , x_{n}$ is given, for each ${\mathcal P}\in {\mathcal F}$, let the  NML distribution associated with ${\mathcal P}$ be\\
\begin{eqnarray}\label{p1}
p_{_{\rm NML}}({\bm x};{\mathcal P})&=&\frac{\max _{p\in {\mathcal P}}p({\bm x})}{\sum _{\bm y} \max_{p'\in {\mathcal P}}p'({\bm y})},\nonumber
\end{eqnarray}\\
The posterior probability distribution of ${\mathcal P}$ for given ${\bm x}$ is given by \\
\begin{eqnarray*}\label{anealing0}
p({\mathcal P}|{\bm x})&=& \frac{(p_{_{\rm NML}}({\bm x};{\mathcal P})p({\mathcal P}))^{\beta}}{\sum _{{\mathcal P'}\in {\mathcal F}}(p_{_{\rm NML}}({\bm x};{\mathcal P}')p({\mathcal P}'))^{\beta}}\\
&=&\frac{\exp (-\beta (L_{_{\rm NML}}({\bm x};{\mathcal P})-\log p({\mathcal P})))}{\sum _{{\mathcal P'}
}\exp (-\beta( L_{_{\rm NML}}({\bm x};{\mathcal P}')-\log p({\mathcal P}')))},
\end{eqnarray*}\\
where $0< \beta \leq 1$ is a temperature parameter.

Thus  Ddim
for model fusion ${\mathcal F}^{\odot }$
associated with ${\bm x}$, which we write as ${\rm Ddim}({\mathcal F}^{\odot }|{\bm x})$, is calculated as\\
\begin{eqnarray}\label{datadim}
{\rm Ddim}({\mathcal F}^{\odot }|{\bm x})
&=&\lim _{n\rightarrow \infty} \frac{\log \sum _{{\mathcal P}\in {\mathcal F}}p({\mathcal P}|{\bm x})\log C_{n}({\mathcal P})}{\log n}\nonumber \\
&\geq &\sum _{{\mathcal P}\in {\mathcal F}}p({\mathcal P}|{\bm x}){\rm Ddim}({\mathcal P}).
\end{eqnarray}\\
We have used the relation in Theorem \ref{fusionddim}.
We call (\ref{datadim}) the {\em pseudo Ddim for model fusion associated with ${\bm x}$}, which we write as
$\underline{{\rm Ddim}}({\mathcal F}^{\odot }|{\bm x})$. To determine the model dimensionality as (\ref{datadim}) from data may be called {\em continuous model selection}.
}
\end{example}

We next consider {\em model concatenation} where a number of model classes are concatenated along the timeline as in Fig.~\ref{fig1} (b).
Let
${\mathcal F}=\{{\mathcal P}_{1},\dots , {\mathcal P}_{s}\}$
be a family of model classes.
Let a set of precision parameters $\{r_{i}>0 :i=1,\dots ,s,\  \sum_{i=1}^{s}r_{i}=1\}$.
For $\epsilon >0$, let $\epsilon _{i}=\epsilon ^{r_{i}}\ (i=1,\dots , s )$. Then
$\epsilon =\prod ^{s}_{i=1}\epsilon _{i}$.
We write
model concatenation over ${\mathcal F}$ with ratio $(r_{1}:\dots :r_{s})$ as
${\mathcal F}^{\otimes}={\mathcal P}_{1}\otimes \cdots \otimes {\mathcal P}_{s}$, which means that a model class ${\mathcal P}_{i}$ is specified with precision $\epsilon _{i}=\epsilon ^{r_{i}}$ for any $\epsilon >0$.
Then
the number of points $m(\epsilon : {\mathcal F}^{\otimes})$  is given by\\
\begin{eqnarray}\label{wa}
\log m(\epsilon :{\mathcal F}^{\otimes })&=&\log m(\epsilon _{1} :{\mathcal P}_{1})+\cdots +\log m (\epsilon _{s} :{\mathcal P}_{s}).
\end{eqnarray}\\
Then Ddim of ${\mathcal F}^{\otimes}$ with ratio $(r_{1}:\dots  :r_{s})$  is calculated as follows:
\begin{eqnarray}\label{ddimconc}
{\rm Ddim}({\mathcal F}^{\otimes})\buildrel \rm def \over =\lim _{_{\small {\begin{array}{c}\forall i, \epsilon _{i} \rightarrow 0\\
\forall i, \frac{\log \epsilon _{i}}{\log \epsilon }=r_{i}={\rm const.}
\end{array}}}}\frac{\log m(\epsilon :{\mathcal F}^{\otimes})}{\log (1/\epsilon )}.
\end{eqnarray}

As for Ddim for model concatenation
we have the following theorem:
\begin{theorem}\label{conc0}
Let ${\mathcal F}=\{{\mathcal P}_{1},\dots , {\mathcal P}_{s}\}$. Ddim for model concatenation of ${\mathcal F}^{\otimes}$ with ratio $(r_{1}:\cdots :r_{s})$ is given by
\begin{eqnarray}\label{pcon}
{\rm Ddim}({\mathcal F}^{\otimes})=\sum _{i=1}^{s}r_{i}{\rm Ddim }({\mathcal P}_{i}). \\ \nonumber
\end{eqnarray}
\end{theorem}
{\em Proof.}
The definition (\ref{ddimconc}) of Ddim of model concatenation with the property (\ref{wa}) directly induces the following:
\begin{eqnarray*}
{\rm Ddim}({\mathcal F}^{\otimes})&=&\lim _{\epsilon \rightarrow 0}\frac{\log m(\epsilon ;{\mathcal F}^{\otimes})}{\log (1/\epsilon )}\\
&=&\lim _{\forall i, \epsilon _{i}\rightarrow 0}\sum _{i=1}^{s}\frac{\log m(\epsilon _{i};{\mathcal P}_{i})}{\frac{1}{r_{i}}\log (1/\epsilon_{i})}\\
&=&\sum _{i=1}^{s}r_{i}{\rm Ddim}({\mathcal P}_{i}).\\
\end{eqnarray*}
This proves (\ref{pcon}).
\hspace*{\fill}$\Box$\\

Combining Theorem \ref{conc0} with Theorem \ref{th1} yields the following corollary.
\begin{corollary}\label{conc1}
Let ${\mathcal F}=\{{\mathcal P}_{1},\dots , {\mathcal P}_{s}\}$ where ${\mathcal P}_{i}$ is the model with parametric  dimensionality $k_{i}$ $(i=1,\dots ,s)$.
Under the condition as in Theorem \ref{basic} for  each ${\mathcal P}_{i}$,
Ddim for model concatenation over ${\mathcal F}^{\otimes }={\mathcal P}_{1}\otimes \cdots \otimes {\mathcal P}_{s}$ with ratio $(r_{1},\dots , r_{s})$ is given as follows:
\begin{eqnarray*}\label{modelconcddim}
{\rm Ddim} ({\mathcal F}^{\otimes })=\sum _{i=1}^{s}r_{i}k_{i}.
\end{eqnarray*}
\end{corollary}

\section{Rate of Convergence of Learning NML Distributions}

This section gives the rates of convergence of the MDL learning algorithm and relates them to Ddim.
It selects a model with the shortest total codelength required for encoding the data as well as the model itself.
It is formalized as follows:

Let ${\mathcal F}=\{{\mathcal P}_{1},\dots , {\mathcal P}_{s}\}$ where $|{\mathcal F}|=s< \infty$.
For a given training data sequence ${\bm x}=x_{1},\dots ,x_{n}$ where each $x_{i}$ is independently drawn, the MDL learning algorithm selects $\hat{{\mathcal P}}$ such that \\
\begin{eqnarray}\label{mdllearning}
\hat{{\mathcal P}}&=&\argmin _{{\mathcal P}\in {\mathcal F}}(-\log p_{_{\rm NML}}({\bm x}; {\mathcal P}))\\
&=&
\argmin_{{\mathcal P}\in {\mathcal F}}\left\{-\log \max _{p\in {\mathcal P}}p({\bm x})+\log {\mathcal C}_{n}({\mathcal P})
\right\}, \nonumber
\end{eqnarray}\\
where ${\mathcal C}_{n}({\mathcal P})$ is the parametric complexity of ${\mathcal P}$ as in (\ref{int1}).
The MDL learning algorithm outputs the NML distribution  associated with $\hat{{\mathcal P}}$ as in (\ref{mdllearning}): For a sequence ${\bm y}=y_{1},\dots , y_{n}$ \\
\begin{eqnarray}\label{mdloutput}
\hat{p}({\bm y})=\frac{\max _{p\in \hat{{\mathcal P}}}p({\bm y})}{C_{n}(\hat{{\mathcal P}})}. \\ \nonumber
\end{eqnarray}
Note that ${\bm y}$ is independent of the training sequence ${\bm x}$ used to obtain $\hat{{\mathcal P}}$.

 We have the following theorem relating Ddim to  the rate of convergence of the  MDL learning algorithm.
\begin{theorem}\label{rate2}
Suppose that each ${\bm x}$ is
generated according to 
$ p^{*}\in {\mathcal P}^{*}\in {\mathcal F}=\{{\mathcal P}_{1},\dots , {\mathcal P}_{s}\}$.
Let $\hat{p}$ be the output of the MDL learning algorithm
as in (\ref{mdloutput}).
Let $d_{B}^{(n)}(\hat{p},p^{*})$ be the Bhattacharyya distance
between $\hat{p}$ and $p^{*}$: \\
\begin{eqnarray}\label{bha}
d_{B}^{(n)}(\hat{p},p^{*})\buildrel \rm def \over =-\frac{1}{n}\log \sum _{{\bm y}} (p^{*}({\bm y})\hat{p}({\bm y}))^{\frac{1}{2}}.
\end{eqnarray}\\
Then for any $\epsilon >0$,
we have the following upper bound on
the probability that
the Bhattacharyya distance between the output of the MDL learning algorithm and the true distribution exceeds $\epsilon$: \\
\begin{eqnarray}\label{000}
Prob[d_{B}^{(n)}(\hat{p},p^{*})>\epsilon ]&<  &\exp \left(-n\epsilon +\frac{1}{2}\log C_{n}({\mathcal P}^{*})+\log |{\mathcal F}|\right).
\end{eqnarray}\\
Further under the condition for ${\mathcal P}^{*}$ as in Theorem \ref{basic} or \ref{basic2}, we have\\
\begin{eqnarray}\label{00111}
Prob[d_{B}^{(n)}(\hat{p},p^{*})>\epsilon ]&=&O\left(n^{{\rm Ddim}({\mathcal P}^{*})/4}e^{-n\epsilon}\right). \\ \nonumber
\end{eqnarray}
\end{theorem}

Theorem \ref{rate2} implies that if $\epsilon >((1/2)\log C_{n}({\mathcal P}^{*})+\log |{\mathcal F}|)/n$, the NML distribution with model of the shortest NML codelength converges exponentially to the true distribution in probability as $n$ increases and the rate is governed by Ddim for the true model.

For $0< \alpha <1$, we define the $\alpha$-divergence between two probability mass functions: $p$ and $q$ by
\begin{eqnarray*}\label{alpha}
& &d^{(n)}_{\alpha}(p,q)\buildrel \rm def \over = \frac{1}{2\alpha (1-\alpha )}\left\{ 1-\left(\sum _{{\bm x}}\{p({\bm x})\}^{\alpha}\{q({\bm x})\}^{1-\alpha}\right)^{\frac{1}{n}}\right\}
\end{eqnarray*}
The Hellinger distance $d_{H}^{(n)}(p, q)$ is defined as a specific case of $\alpha =1/2$: $d_{H}^{(n)}(p,q)=d_{1/2}^{(n)}(p,q)$. Since we can easily verify the following relation between the Hellinger distance and the Bhacchataryya distance:
\[d_{H}^{(n)}(p,q)\leq 2d_{B}^{(n)}(p,q),\]
the results (\ref{000}) and (\ref{00111}) on the rates of convergence hold for the Hellinger distance, within a constant factor.
Further, using a similar technique of the proof of Theorem \ref{rate2} as shown below, we can verify that the same results hold for the $\alpha$ divergence with  $0< \alpha <1$, within a constant factor.

\ \ \\
{\em Proof of Theorem \ref{rate2}.}
Let $p^{*}$ be the true distribution 
 associated with the true model ${\mathcal P}^{*}$.
Let $\hat{{\mathcal P}}$ be the model selected by the MDL learning algorithm and let
$p_{_{\rm NML}}({\bm x};\hat{{\mathcal P})}$ be the NML distribution associated with $\hat{{\mathcal P}}$. We write it as $\hat{p}$.

By the definition of the MDL learning algorithm,
we have\\
\begin{eqnarray}\label{in}
\min_{{\mathcal P}}( -\log p_{_{\rm NML}}({\bm x}; {\mathcal P}))
& \leq &-\log p_{_{\rm NML}}({\bm x}; {\mathcal P}^{*})\nonumber \\
& =&-\log \max _{p\in {\mathcal P}^{*}}p({\bm x})+\log C_{n}({\mathcal P}^{*})\nonumber \\
& \leq &
-\log p^{*}({\bm x})+\log C_{n}({\mathcal P}^{*}). \\ \nonumber
\end{eqnarray}

Let $p_{_{\rm NML},{\mathcal P}}$ be the NML distribution $p_{_{\rm NML}}({\bm x}:{\mathcal P})$ associated with ${\mathcal P}$ defined as\\
\[p_{_{\rm NML}}({\bm x}:{\mathcal P})=\frac{\max _{p\in {\mathcal P}}p({\bm x})}{C_{n}({\mathcal P})}.\\ \]
For $\epsilon >0$, the following inequalities hold:\\
\begin{align}
&Prob[d_{B}^{(n)}(\hat{p},p^{*})>\epsilon ] \nonumber \\
&\leq
Prob[ {\bm x}: \ (\ref{in})\ {\rm holds\ under}\ d_{B}^{(n)}(\hat{p},p^{*})>\epsilon ] \nonumber \\
&= Prob\left[ {\bm x}: \min _{{\mathcal P}:d_{B}^{n}(p_{_{\rm NML},{\mathcal P}},p^{*})>\epsilon }(-\log p_{_{\rm NML}}({\bm x};{\mathcal P}))\leq -\log p^{*}({\bm x})+\log C_{n}({\mathcal P}^{*})\right] \nonumber \\
&=Prob\left[ {\bm x}: \max _{{\mathcal P}:d_{B}^{n}(p_{_{\rm NML},{\mathcal P}},p^{*})>\epsilon }p_{_{\rm NML}}({\bm x}:{\mathcal P})\geq p^{*}({\bm x})/C_{n}({\mathcal P}^{*})\right]\nonumber \\
&\leq \sum _{{\mathcal P}\in {\mathcal F}, d_{B}^{(n)}(p_{_{\rm NML},{\mathcal P}}, p^{*})>\epsilon}Prob\bigl[ {\bm x}:
p_{_{\rm NML}}({\bm x}:{\mathcal P})\geq p^{*}({\bm x})/C_{n}({\mathcal P}^{*}) \bigr]. \label{sumevent00}\\
\nonumber 
\end{align}

Let $E_{n}({\mathcal P})$ be the event that\\
\begin{eqnarray*}
p_{_{\rm NML}}({\bm x}:{\mathcal P})
\geq p^{*}({\bm x})/C_{n}({\mathcal P}^{*}).\\
\end{eqnarray*}

Note that under the event $E_{n}({\mathcal P})$,
we have \\
\[1\leq \left(\frac{p_{_{\rm NML}}({\bm x};{\mathcal P})}{p^{*}({\bm x})}\right)^{\frac{1}{2}} ( C_{n}({\mathcal P}^{*}))^{\frac{1}{2}}. \]\\
Then under the condition that $d_{B}^{(n)}(p_{_{\rm NML},{\mathcal P}}, p^{*})>\epsilon$, we have\\
\begin{eqnarray}
Prob[E_{n}({\mathcal P})]&=&\sum _{{\bm x}\cdots E_{n}({\mathcal P})}p^{*}({\bm x})\nonumber \\
&\leq &\sum _{{\bm x}\cdots E_{n}({\mathcal P})}p^{*}({\bm x})\left(\frac{p_{_{\rm NML}}({\bm x}; {\mathcal P})}{p^{*}({\bm x})}\right)^{\frac{1}{2}} ( C_{n}({\mathcal P}^{*}))^{\frac{1}{2}} \nonumber \\
&\leq& \left\{\sum_{{\bm y}} (p_{_{\rm NML}}({\bm y};{\mathcal P})p^{*}({\bm y}))^{\frac{1}{2}} \right\} (C_{n}({\mathcal P}^{*}))^{\frac{1}{2}} \nonumber \\
&< &\exp (-n\epsilon+(\log C_{n}({\mathcal P}^{*}))/2) ,
\label{event100}
\end{eqnarray}\\
where  we have used the fact that by (\ref{bha}), under $d_{B}^{(n)}(p_{_{\rm NML},{\mathcal P}},p^{*})>\epsilon$, it holds \\
\[\sum _{{\bm y}} (p_{_{\rm NML}}({\bm y};{\mathcal P})p^{*}({\bm y}))^{\frac{1}{2}}< e^{-n\epsilon }.\]\\
Plugging (\ref{event100}) into (\ref{sumevent00}) yields\\
\begin{eqnarray*}
Prob[d_{B}^{(n)}(\hat{p},p^{*})>\epsilon ]
&\leq &\sum _{{\mathcal P}\in {\mathcal F}, d_{B}^{(n)}(p_{_{\rm NML},{\mathcal P}}, p^{*})>\epsilon}Prob[E_{n}({\mathcal P})]\\
&< &\sum _{{\mathcal P}\in {\mathcal F}, d_{B}^{(n)}(p_{_{\rm NML},{\mathcal P}}, p^{*})>\epsilon}\exp \left(-n\epsilon +(1/2)\log C_{n}({\mathcal P}^{*})\right)\\
&\leq &\sum _{{\mathcal P}\in {\mathcal F}} \exp \left(-n\epsilon +(1/2)\log C_{n}({\mathcal P}^{*})\right)\\
&=&
\exp \left(-n\epsilon +(1/2)\log C_{n}({\mathcal P}^{*})+\log |{\mathcal F}|\right).
\end{eqnarray*}\\
This implies (\ref{000}). Further note that under the condition for ${\mathcal P}^{*}$ as in Theorem \ref{basic} or Theorem \ref{basic2}, the following  asymptotic relation holds:\\
\begin{eqnarray}\label{mk}
\frac{1}{2}\log C_{n}({\mathcal P}^{*})+\log |{\mathcal F}|=\frac{1}{4}{\rm Ddim}({\mathcal P}^{*})\log n+o(\log n).\\ \nonumber
\end{eqnarray}
Plugging (\ref{mk}) into (\ref{000}) yields (\ref{00111}).
This completes the proof.
\hspace*{\fill}$\Box$\\

Theorem \ref{rate2} has dealt with
the case where the true distribution $p^{*}$ is in some ${\mathcal P}\in {\mathcal F}$.
We may be further interested in the {\em agnostic case} where the true distribution $p^{*}$ is not necessarily in some ${\mathcal P}\in {\mathcal F}$.
Theorem \ref{rate2n} shows the rate of convergence of the MDL learning algorithm for such an agnostic case.
\begin{theorem}\label{rate2n}
Let $p^{*}$ be the true distribution and $\hat{p}$ be the output of the MDL learning algorithm,
 Let $D(p^{*}||p)\buildrel \rm def \over =\lim _{n \rightarrow \infty}(1/n)\sum _{{\bm x}}p^{*}({\bm x})\log (p^{*}({\bm x})/p({\bm x}))$ be the Kullback-Leibler divergence between $p^{*}$ and $p$.
For $\epsilon >0$, let $A_{n,\epsilon}$ be the event that for any ${\mathcal P}\in {\mathcal F}$, for $\tilde{p}=\argmin _{p\in {\mathcal P}}D(p^{*}||p),$
\begin{eqnarray}\label{exevent}
 \left|D(p^{*}||\tilde{p})-\frac{1}{n}\log \frac{ p^{*}({\bm x})}{\tilde{p}({\bm x})}\right|<\epsilon  .\\ \nonumber
 \end{eqnarray}
 Let $P_{n,\epsilon}\buildrel \rm def \over =Prob[A_{n,\epsilon}^{c}]$
 where $A_{n,\epsilon}^{c}$ is the complementary set of $A_{n,\epsilon}$. Then for any $\epsilon >0$, the probability that the Bhattacharyya distance between $p^{*}$ and $\hat{p}$,  is upper-bounded as follows:\\
\begin{eqnarray}\label{agnosticbound}
Prob[d^{(n)}
_{B}(\hat{p},p^{*})>\epsilon ]&< &\frac{|{\mathcal F}|}{1-P_{n,\epsilon}}\exp \left( -\frac{n}{2}\left( \epsilon -\frac{J_{n}(p^{*})}{n}\right)\right)+P_{n,\epsilon},
\\ \nonumber
\end{eqnarray}
where
\begin{eqnarray*}
J_{n}(p^{*})\buildrel \rm def \over =\min _{{\mathcal P}}\left\{n\inf _{p\in {\mathcal P}}D(p^{*}||p)+\log C_{n}({\mathcal P})\right\}.\\
\end{eqnarray*}
Specifically,
if for some function $B_{n}$ of $n$,  for any ${\mathcal P}$,
$(1/n)|\log ( p^{*}({\bm x})/\tilde{p}({\bm x})|\leq B_{n}$, then we have\\
\begin{eqnarray*}
Prob[A_{n, \epsilon}^{c}]&\leq &2|{\mathcal F}|\exp \left( -\frac{n\epsilon ^{2}}{2B_{n}^{2}}\right).\\
\end{eqnarray*}
\end{theorem}

Basically, Theorem \ref{rate2n} can be proven similarly with Theorem \ref{rate2}.
However, the bound (\ref{agnosticbound}) in Theorem \ref{rate2} cannot be obtained as a specific case of Theorem \ref{rate2n} where $p^{*}$ is in some ${\mathcal P}^{*}$.
This is due to a technical reason that the convergence of the empirical log likelihood ratios to the KL-divergence  should be explicitly evaluated in the proof of Theorem \ref{rate2n}  while they need not be evaluated in the proof of Theorem \ref{rate2}.
From this reason, we leave the proof of Theorem \ref{rate2} here and move that of Theorem \ref{rate2n} to Appendix A.2.

Theorem \ref{rate2n} shows that the NML distribution with model of the shortest NML codelength converges to the true model in probability as $n$ increases, provided that $\epsilon >(J_{n}(p^{*})+\log |{\mathcal F}|)/n$ and $P_{n,\epsilon}\rightarrow 0$.

Next we consider model fusion where ${\mathcal P}$ is chosen randomly according to the probability distribution $\pi ({\mathcal P})$ over ${\mathcal F}=\{{\mathcal P}_1,\dots , {\mathcal P}_{s}\}$.
Then the unknown
true distribution $p^{*}$ is chosen from ${\mathcal P}^{*}$.
 We have the following corollary relating Ddim for model fusion to  the rate of convergence of the  MDL learning algorithm.

 \begin{corollary}\label{rate3}
 Let $\hat{{\mathcal P}}\in {\mathcal F}$ be the model selected by the MDL learning algorithm and $\hat{p}$ be the NML distribution associated with $\hat{{\mathcal P}}$.
Then under the condition for each ${\mathcal P}\in {\mathcal P}$ as in Theorem \ref{basic} or Theorem \ref{basic2}, for model fusion over ${\mathcal F}$, we have the following upper bound on
the expected Bhattacharyya distance between the output of the MDL learning algorithm and the true distribution: \\
\begin{eqnarray}
E_{{\mathcal P}^{*}}E_{{\bm x}\sim p^{*}\in {\mathcal P}^{*}}[d_{B}^{(n)}(\hat{p},p^{*}) ]
\label{00001}
&=&O\left(\frac{{\rm Ddim}({\mathcal F}^{\odot})\log n}{n}\right). \\ \nonumber
\end{eqnarray}
\end{corollary}

Corollary \ref{rate3} shows that the expected Bhattacharyya distance between the true distribution and the output of the MDL learning algorithm is characterized by Ddim for model fusion over ${\mathcal F}$.\\
\ \ \\
{\em Proof of Corollary \ref{rate3}.}
Let $r_{n}({\mathcal P}^{*})\buildrel \rm def \over =\{(1/2)\log {\mathcal C}_{n}({\mathcal P}^{*})+\log |{\mathcal F}|\}/n$.
For fixed ${\mathcal P}^{*}$, the expected Bhattacharyya distance 
 is given by \\
 \begin{eqnarray}
E_{{\bm x}\sim p^{*}\in {\mathcal P}^{*}}[d_{B}^{(n)}(\hat{p},p^{*})-r_{n}({\mathcal P}^{*})]&=&\int ^{\infty}_{0}Prob[d_{B}^{(n)}(\hat{p},p^{*})-r_{n}({\mathcal P}^{*})>\epsilon ]d\epsilon \nonumber \\
&=&\int ^{\infty}_{0}Prob[d_{B}^{(n)}(\hat{p},p^{*})>\epsilon +r_{n}({\mathcal P}^{*})]d\epsilon \nonumber \\
&\leq &\int ^{\infty}_{0}e^{-n\epsilon}d\epsilon \label{km} \\
&=&\frac{1}{n}, \nonumber \\
\end{eqnarray}
where we have used (\ref{000}) to derive (\ref{km}).
Thus we have\\
\begin{eqnarray*}
E_{{\bm x}\sim p^{*}\in {\mathcal P}^{*}}[d_{B}^{(n)}(\hat{p},p^{*})]&\leq&r_{n}({\mathcal P}^{*})+\frac{1}{n}. \\
 \end{eqnarray*}
By further taking the expectation with respect to ${\mathcal P}^{*}$, we have\\
\begin{eqnarray}
E_{{\mathcal P}^{*}}[d_{B}^{(n)}(\hat{p},p^{*})]
&\leq &E_{{\mathcal P}^{*}}E_{{\bm x}\sim p^{*}\in {\mathcal P}^{*}}[r_{n}({\mathcal P}^{*})]+\frac{1}{n} \nonumber \\
& =& \frac{E_{{\mathcal P}^{*}}[\log C_{n}({\mathcal P}^{*})]}{2n}+\frac{\log |{\mathcal F}|+1}{n} \nonumber \\
&\leq & \frac{\log E_{{\mathcal P}^{*}}[C_{n}({\mathcal P}^{*})]}{2n} +\frac{\log |{\mathcal F}|+1}{n} \label{eq1}\\
&=&O\left(\frac{\log E_{{\mathcal P}^{*}}[m_{n}({\mathcal P}^{*})]}{n}\right) \label{eq2}\\
&=&O\left(\frac{{\rm Ddim} ({\mathcal F}^{\odot})\log n}{n}\right).\label{eq3}\\ \nonumber
 \end{eqnarray}
We have used the Jensen's inequality to derive (\ref{eq1}).
To derive (\ref{eq2}), we have used the fact $\log C_{n}({\mathcal P}^{*})=\log m_{n}({\mathcal P}^{*})+O(1)$
under the conditions in Corollary \ref{rate3}.
Eq.(\ref{eq3}) comes from the definition (\ref{fusiondef}) of Ddim for model fusion. Thus (\ref{00001}) is obtained.
\hspace*{\fill}$\Box$\\

Theorem \ref{rate2} implies that the
rate of convergence of the expected Bhattacharyya distance for  MDL learning is governed by the parametric complexity for the true model, or eventually Ddim for it.
This corresponds to the fact that the rate of convergence for the empirical risk minimization algorithm is governed by
the metric dimension or VC dimension of the target function class~(see e.g. \cite{dudley}, \cite{pollard}, \cite{Haussler}).

In order to compare the result in Theorem \ref{rate2} with existing ones, we describe the previous results on MDL learning according to \cite{barron}, \cite{yamanishi92}.
Let $\overline{\mathcal P}$ be the subset of ${\mathcal P}$ obtained by discretizing ${\mathcal P}$.
We define the {\em two-stage MDL learning algorithm} as the algorithm which takes ${\bm x}$ as input and outputs $\hat{p}\in \overline{\hat{\mathcal P}}$ by selecting the  model $\hat{{\mathcal P}}$ that attains the shortest codelength as follows:
For $\lambda \geq 2$, \\
\begin{eqnarray}
\hat{{\mathcal P}}&=&\argmin _{{\mathcal P}\in {\mathcal  F}}\left\{-\log \max
 _{p\in \overline{{\mathcal P}}\subset {\mathcal P}}p({\bm x})+\lambda \ell (p, {\mathcal P})\right\}, \label{2stage} \\
 \hat{p}&=& \arg \max _{p\in \overline{\hat{{\mathcal P}}}\subset \hat{{\mathcal P}}}p({\bm x}), \label{2stageoutput}
 \end{eqnarray}\\
 where $\ell (p, {\mathcal P})$ is a non-negative valued function satisfying the {\em Kraft's inequality}:\\
 \begin{eqnarray*}
 \sum _{{\mathcal P}\in {\mathcal F}}\sum _{p\in \overline{{\mathcal P}}\subset {\mathcal P}}e^{-\ell (p, {\mathcal P})}\leq 	1.
 \end{eqnarray*}\\
 The righthand side of (\ref{2stage}) is called the {\em two-stage codelength} in the sense that a given data is encoded with two steps; first  the model ${\mathcal P}$ and $p$ are encoded and then the data is encoded on the basis of them.
 We call (\ref{2stageoutput}) the {\em two-stage MDL distribution.}
 Note that the two-stage MDL distribution is obtained over the discretized model class $\overline{{\mathcal P}}.$

Assume that the data is independently identically distributed according to the true distribution $p^{*}$.
We define the Hellinger distance between $\hat{p}$ and $p^{*}$ as $d_{H}(\hat{p},p^{*})\buildrel \rm def \over =\sum _{x}(\sqrt{\hat{p}(x)}-\sqrt{p^{*}(x)})^{2}$. Then according to \cite{barron},
the expected Hellinger distance is given by\\
\begin{eqnarray}\label{resolvability}
E^{n}_{p^{*}}[d_{H}(\hat{p},p^{*})]=O\left( \min _{{\mathcal P}}\inf _{p\in \overline{{\mathcal P}}\subset {\mathcal P}}\left(D(p^{*}||p)+\frac{\ell (p,{\mathcal P})}{n} \right)\right).\\ \nonumber
\end{eqnarray}
The righthand side is called the index of resolvability and the $D(p^{*}||p)$ denotes the Kullback-Leibler divergence between $p^{*}$ and $p$: $D(p^{*}||p)\buildrel \rm def \over =\sum _{x}p^{*}(x)\log (p^{*}(x)/p(x))$.
In the specific case where $p^{*}$ is in ${\mathcal P}^{*}$, we have\\
\begin{eqnarray}\label{ir2}
E^{n}_{p^{*}}[d_{H}(\hat{p},p^{*})]=O\left( \frac{\ell (\tilde{p}^{*},{\mathcal P}^{*})}{n} \right),\\ \nonumber
\end{eqnarray}
where $\tilde{p}^{*}$ is the truncation of $p^{*}$ in $\overline{{\mathcal P}^{*}}\subset {\mathcal P}^{*}$.
 Note that the rates of convergence in (\ref{resolvability}) and (\ref{ir2}) depend on $\ell (p, {\mathcal P})$, which also depends on the method for discretizing ${\mathcal P}$.
 Further note that the Bhattacharyya distance dealt with in Theorems \ref{rate2} and \ref{rate2n} measures the distance between the estimated NML distribution for $n$ sample and the true distribution, where the distance is normalized with respect to sample size. Meanwhile the Hellinger distance in (\ref{ir2})
 measures the distance between the learned distribution for a single sample and the true one.

Theorem \ref{rate2} differs from the conventional results in \cite{barron}, \cite{yamanishi92}
on MDL learning in the following regards:
This paper analyzes the convergence of the NML distribution with model of the shortest NML codelength,
 while previous work analyzed the convergence of the two-stage MDL distribution.
 In the latter the model class should be properly discretized
 in MDL learning where  it is a critical issue how finely we should discretize the parameter space.
Meanwhile, in our analysis any discretization is not required, but rather only the NML codelength is calculated.
Although the algorithm designed without using discretization have  been proposed by Chatterjee and Barron~\cite{cb} and Kawakita and Takeuchi~\cite{kawakita}, their targets of analysis were still the two-stage MDL distribution, for which the discretization used to be an intermediate key technique for analysis.
Note that the NML distribution is the only one that has the following nice properties:
I) it attains Shtarkov's minimax regret, and II) it has estimation optimality \cite{rissanen} (p.57-58).
Hence it is  worthwhile analyzing the convergence of the NML distribution rather than the two-stage MDL distribution.


\section{Error Exponents for MDL-based Model Change Detection}

This section addresses the issue of model change detection.
We relate Dim to the performance of MDL-based model change detection.
Let ${\mathcal F}$ be a family of model classes.
Let ${\bm x}=x_{1},\dots ,x_{n}$ be an observed sequence where each $x_{i}$ is independently drawn.
For a given positive integer $m$, let ${\bm t}=(t_{1},\dots , t_{m}),\ (1<t_{1}<t_{2}<\cdots < t_{m}<n)$ be a sequence of change points and ${\bm {\mathcal P}}$ be a sequence of models: ${\mathcal P}_{(0)},\dots ,{\mathcal P}_{(m)}$ where ${\mathcal P}_{(i)}\in {\mathcal F}\ (i=1,\dots ,m)$
and ${\mathcal P}_{(i-1)}\neq {\mathcal P}_{(i)}\ (i=1,\dots , m)$.
Suppose that ${\bm x}$ is generated according to
a series of the following probability distributions:\\
\begin{eqnarray*}
x_{1}^{t_{1}}&\sim &\exists p_{0}({\bm x})\in {\mathcal P}_{(0)},\\
x_{t_{1}+1}^{t_{2}}&\sim &\exists p_{1}({\bm x})\in {\mathcal P}_{(1)},\\
\cdots & &\cdots\\
x_{t_{m}+1}^{n}&\sim & \exists p_{m}({\bm x})\in {\mathcal P}_{(m)}.\\
\end{eqnarray*}
We say that ${\bm x}$ is generated from a triplet: $\{m, {\bm t},  {\bm {\mathcal P}}\}$ where $m$ is the number of change points, ${\bm t}$ is the sequence of change point locations, and ${\bm {\mathcal P}}$ is a model sequence.
Below we suppose that ${\bm x}$ is generated from some $\{m, {\bm t},  {\bm {\mathcal P}}\}$.

We define the codelength ${\mathcal L}({\bm x}; {\bm {\mathcal P}})$ as\\
\begin{eqnarray*}
{\mathcal L}({\bm x}; {\bm {\mathcal P}})&\buildrel \rm def \over =&\sum ^{m}_{i=0}L_{_{\rm NML}}(x^{t_{i-1}}_{t_{i}-1}; {\mathcal P}_{(i)})\\ 
&=&
\sum ^{m}_{i=0}\left\{-\log \max _{p\in {\mathcal P}_{(i)}}p(x^{t_{i-1}}_{t_{i}-1})+\log C_{t_{i}-t_{i-1}+1} ({\mathcal P}_{(i)})\right\}. \\
\end{eqnarray*}

For a given triplet $\{m^{*}, {\bm t}^{*},  {\bm {\mathcal P}}^{*}\}$ (${\bm t}^{*}=(t^{*}_1,\dots , t_{m}^{*}), {\bm {\mathcal P}}^{*}={\mathcal P}_{(0)}^{*},\dots ,{\mathcal P}_{(m^{*})}^{*}),$ we conduct a hypothesis testing whether ${\bm x}$ is generated from $\{m^{*}, {\bm t}^{*},  {\bm {\mathcal P}}^{*}\}$  or not.
Intuitively, if there exists some  triplet  that compresses the data significantly more than  $\{m^{*}, {\bm t}^{*},  {\bm {\mathcal P}}^{*}\}$, we determine that the data is not generated from it, otherwise we determine that the data is generated from $\{m^{*}, {\bm t}^{*},  {\bm {\mathcal P}}^{*}\}$.
We define the {\em MDL change statistics} as follows:\\
\begin{eqnarray}\label{mdlchangestatisticsn}
\Phi ({\bm x})\buildrel \rm def \over ={\mathcal L}({\bm x}; {\bm {\mathcal P}}^{*})-\min _{{\bm {\mathcal P}}}\{{\mathcal L}({\bm x};{\bm {\mathcal P}})+\ell ({\bm {\mathcal P}})\}-n\epsilon ,\\ \nonumber
\end{eqnarray}
where $\ell ({\bm {\mathcal P}})(>0)$ is the function satisfying Kraft's inequality:
$\sum _{{\bm {\mathcal P}}\in {\mathcal F}^{m}}e^{-\ell ({\bm {\mathcal P}})}\leq 1$.
We define the {\em MDL test} for $\{m^{*}, {\bm t}^{*},  {\bm {\mathcal P}}^{*}\}$ as:
If $\phi ({\bm x})>0$, then we accept ${\mathcal H}_{0}$: ${\bm x}$ is generated from $\{m^{*}, {\bm t}^{*},  {\bm {\mathcal P}}^{*}\}$ otherwise we accept ${\mathcal H}_{1}$:  ${\bm x}$ is not generated from it but rather is generated from another unknown $\{ \tilde{m}, \tilde{{\bm t}},  \tilde{{\bm {\mathcal P}}}\}$ where ($\tilde{{\bm t}}=\tilde{t}_1,\dots , \tilde{t}_{\tilde{m}}), \tilde{{\bm {\mathcal P}}}=\tilde{{\mathcal P}}_{(0)},\dots ,\tilde{{\mathcal P}}_{(\tilde{m})}$.

As performance metrics,
we define {\em Type I error probability} as the probability that
${\mathcal H}_{1}$ is accepted while the true hypothesis is ${\mathcal H}_{0}$.
We define {\em Type II error probability} as the probability that
${\mathcal H}_{0}$ is accepted while the true hypothesis is ${\mathcal H}_{1}$.

Theorem \ref{epn} shows Type I and II error probabilities for the MDL test.
\begin{theorem}\label{epn}{
Under the conditions  as in Theorem \ref{basic} or \ref{basic2} for any ${\mathcal P}\in {\mathcal F}$,
For any $\epsilon >0$, Type I error probability for the MDL test is 
upper-bounded by\\
\begin{eqnarray}\label{type1n}
n^{2m^{*}{\rm Ddim}({\mathcal P}_{(0)}^{*}\otimes \cdots \otimes {\mathcal P}_{(m^{*})}^{*})(1+o(1))}\exp (-n\epsilon),
\end{eqnarray}\\
where ${\mathcal P}_{(0)}^{*}\otimes \cdots \otimes {\mathcal P}_{(m^{*})}^{*}$ is model concatenation of ${\mathcal P}_{(0)}^{*},\dots ,{\mathcal P}_{(m^{*})}^{*}$ with ratio
$(\log t_{1}^{*}:\log (t_{2}^{*}-t^{*}_{1}):\cdots : \log (n-t^{*}_{m^{*}})$.
Type II error probability for the MDL test is 
upper-bounded by \\
\begin{eqnarray}\label{type2n}
n^{\tilde{m}{\rm Ddim}(\tilde{{\mathcal P}}_{(0)}\otimes\cdots \otimes \tilde{{\mathcal P}}_{(\tilde{m})})(1+o(1))}
\exp (-n(d_{B}^{n}
({\bm {\mathcal P}}^{*}_{{\rm NML}}; {\tilde{\bm {\mathcal P}}})-\ell (\tilde{\bm {\mathcal P}})/2n-\epsilon /2)),  \\ \nonumber
\end{eqnarray}
where $\tilde{{\mathcal P}}_{(0)}\otimes\cdots \otimes \tilde{{\mathcal P}}_{(\tilde{m})}$ is model concatenation of
$\tilde{{\mathcal P}}_{(0)}, \cdots ,  \tilde{{\mathcal P}}_{(\tilde{m})}$ with ratio  $(\log  \tilde{t}_{1}: \log (\tilde{t}_{2}-\tilde{t}_{1}):\cdots :\log (n-\tilde{t}_{m}))$.
$d_{B}^{n}
({\bm {\mathcal P}}^{*}_{_{\rm NML}}; \tilde{\bm {\mathcal P}})$ is
the Bhattacharyya distance between the concatenated NML distribution associated with
${\bm {\mathcal P}}^{*}={\mathcal P}_{(0)}^{*}\otimes \cdots \otimes  {\mathcal P}^{*}_{(m^{*})}$ and the concatenated distribution $\prod ^{\tilde{m}}_{j=0}\tilde{p}\left(x^{\tilde{t}_{j+1}}_{\tilde{t}_{j}+1}\right)$ assiciated with the true model sequence:
$\tilde{\bm {\mathcal P}}=\tilde{P}_{(0)}\otimes \cdots \otimes \tilde{{\mathcal P}}_{(\tilde{m})}$: \\
\begin{eqnarray*}d_{B}^{(n)}({\bm {\mathcal P}}^{*}_{_{\rm NML}}; \tilde{\bm {\mathcal P}})\buildrel \rm def \over =-\frac{1}{n}\log \sum _{{\bm x}} \left\{\left(\prod ^{m^{*}}_{i=0}p_{_{\rm NML}}\left(x^{t^{*}_{i+1}}_{t^{*}_{i}+1};{\mathcal P}^{*}_{(i)}\right)\right) \left(\prod ^{\tilde{m}}_{j=0}\tilde{p}\left(x^{\tilde{t}_{j+1}}_{\tilde{t}_{j}+1}\right)\right)\right\}^{\frac{1}{2}}.\\
\end{eqnarray*}
}
\end{theorem}

Theorem \ref{epn} implies that  for any $\epsilon$ larger than some value, Type I and II error probabilities for the MDL test decay exponentially to zero as $n$ increases, and are governed by Ddim for model concatenation of the true models  (see (\ref{type1n}) and (\ref{type2n})).

\ \ \\
{\em Proof of Theorem \ref{epn}.}
To evaluate Type I error probability, assume that the MDL change statistics (\ref{mdlchangestatisticsn}) satisfies: $\Phi ({\bm x})>0$. Then for the true model ${\bm {\mathcal P}}^{*}$,
\\
\begin{eqnarray*}
0< \Phi ({\bm x})&
= &
\sum ^{m^{*}}_{i=0}\left\{-\log \max _{p\in {\mathcal P}^{*}_{(i)}}p(x^{t^{*}_{i-1}}_{t^{*}_{i}-1})+\log C_{t^{*}_{i}-t^{*}_{i-1}+1} ({\mathcal P}^{*}_{(i)})\right\}
\\
& &-\min _{{\bm {\mathcal P}}}\left\{{\mathcal L}({\bm x}; {\bm {\mathcal P}})+\ell ({\bm {\mathcal P}})\right\}-n\epsilon
\\
&\leq &\sum ^{m^{*}}_{i=0}\left\{-\log p(x^{t^{*}_{i-1}}_{t^{*}_{i}-1}:{\mathcal P}_{i}^{*})+\log C_{t^{*}_{i}-t^{*}_{i-1}+1} ({\mathcal P}^{*}_{(i)})\right\}
\\
& &-\min _{{\bm {\mathcal P}}}\left\{{\mathcal L}({\bm x}; {\bm {\mathcal P}})+\ell ({\bm {\mathcal P}})\right\}-n\epsilon , \\
\end{eqnarray*}
where $p(x^{t^{*}_{i-1}}_{t^{*}_{i}-1}:{\mathcal P}_{i}^{*})$ is the true probability distribution for the $i+1$th segment $(i=1,\dots ,m^{*})$.
Letting $p({\bm x};{\bm {\mathcal P}}^{*})\buildrel \rm def \over =\prod ^{m^{*}}_{i=0} p(x^{t^{*}_{i-1}}_{t^{*}_{i}-1}:{\mathcal P}_{i}^{*})$, we rewrite this inequality as follows:
\begin{eqnarray*}
p({\bm x}; {\bm {\mathcal P}}^{*})
&\leq&
\exp \left(-\min _{{\bm {\mathcal P}}}({\mathcal L}({\bm x};  {\bm {\mathcal P}})+\ell ({\bm {\mathcal P}}))\right)\\
& & \times \exp \left( -n\epsilon +\sum _{i=0}^{m^{*}}\log C_{t^{*}_{i}-t^{*}_{i-1}+1} ({\mathcal P}^{*}_{(i)})
\right).
\end{eqnarray*}\\
Then Type I error probability is evaluated as follows:\\
\begin{eqnarray}
\sum _{ {\bm x}:
\Phi ({\bm x})>0
}p({\bm x}; {\bm {\mathcal P}}^{*})
&\leq &\sum _{ {\bm x}}\exp \left(-\min _{{\bm {\mathcal P}}}\left({\mathcal L}({\bm x};  {\bm {\mathcal P}})+\ell ({\bm {\mathcal P}})\right)\right)
\label{e0n}\\
& & \times \exp \left( -n\epsilon +\sum _{i=0}^{m^{*}}\log C_{t^{*}_{i}-t^{*}_{i-1}+1} ({\mathcal P}^{*}_{(i)})
\right) \nonumber \\
&\leq &\exp \left( -n\epsilon +\sum _{i=0}^{m^{*}}\log C_{t^{*}_{i}-t^{*}_{i-1}+1} ({\mathcal P}^{*}_{(i)})
\right),\label{e000n}
\\ \nonumber
\end{eqnarray}
where we used the Kraft's inequality, which makes the first term in (\ref{e0n}) not more than $1$ for the prefix codelength.
This is because ${\bm x}$ can be encoded in the two-stages, once the model $\hat{{\mathcal P}}=\argmin _{{\bm {\mathcal P}}}\left({\mathcal L}({\bm x};  {\bm {\mathcal P}})+\ell ({\bm {\mathcal P}})\right)$ is given.
Note that under the assumption in Theorem
\ref{ep} for each ${\mathcal P}\in {\mathcal F}$,
letting $T(n,{m}^{*},{\bm t}^{*})\buildrel\rm def \over =
\log \prod _{i=1}^{{m}^{*}}({t}^{*}_{i}-{t}^{*}_{i-1}+1) $, we have
\begin{align}
&\sum ^{{m}^{*}}_{i=0}\left(
\log C_{{t}^{*}_{i}-{t}^{*}_{i-1}+1} ({{\mathcal P}^{*}}_{(i)})\right) \nonumber \\
&=T(n,{m}^{*},{\bm  t}^{*})\times
\sum ^{{m}^{*}}_{i=0}\frac{\log ({t}^{*}_{i}-{t}^{*}_{i-1}+1)}{T(n, {m}^{*},{\bm t}^{*})}\times \frac{\log C_{{t}^{*}_{i}-{t}^{*}_{i-1}+1}({P}^{*}_{(i)})}{\log ({t}^{*}_{i}-{t}^{*}_{i-1}+1)
} \nonumber \\
&\leq 2{m}^{*}\log (n/{m}^{*}){\rm Ddim}({P}^{*}_{(0)}\otimes \cdots \otimes {P}^{*}_{({m}^{*})})(1+o(1)). \label{chi} \\ \nonumber
\end{align}
Combining (\ref{chi}) with (\ref{e000n}) yields (\ref{type1n}).

To evaluate Type II error probability, assume
that $\Phi ({\bm x})\leq 0$.
Let $\tilde{\bm t}=(\tilde{t}_{0},\dots , \tilde{t}_{\tilde{m}})$. be the change point sequence for ${\mathcal H}_{1}$ and
$\tilde{p}(x^{\tilde{t}_{j+1}}_{\tilde{t}_{j}+1})$ be the true probability distribution for the $(j+1)$th segment $(j=0,\dots ,\tilde{m)}$.
Then for the true model $
p({\bm x};\tilde{{\bm {\mathcal P}}} )\buildrel \rm def \over =
\prod ^{\tilde{m}}_{j=0}\tilde{p}\left(x^{\tilde{t}_{j+1}}_{\tilde{t}_{j}+1}\right)$, it holds:\\
\begin{eqnarray*}
0&\geq &\Phi ({\bm x})\\
&=&
{\mathcal L}({\bm x}; {\bm {\mathcal P}}^{*})-\min _{{\bm {\mathcal P}}}\left\{\sum ^{m}_{i=0}\left(
-\log \max _{p\in {\mathcal P}_{(i)}}p(x^{t_{i-1}}_{t_{i}-1})+\log C_{t_{i}-t_{i-1}+1} ({\mathcal P}_{(i)})\right)
+\ell ({\bm {\mathcal P}})\right\}-n\epsilon \\
&\geq &{\mathcal L}({\bm x}; {\bm {\mathcal P}}^{*})-\left\{\sum ^{\tilde{m}}_{j=0}\left(
-\log p({\bm x}; \tilde{\bm {\mathcal P}}) 
+\log C_{\tilde{t}_{j}-\tilde{t}_{j-1}+1} (\tilde{{\mathcal P}}_{(j)})\right)
+\ell (\tilde{{\bm {\mathcal P}}})\right\}-n\epsilon .\\
\end{eqnarray*}
Letting $p_{_{\rm NML}}
({\bm x}; {\bm {\mathcal P}}^{*})\buildrel \rm def \over =\prod ^{m^{*}}_{i=0}p_{_{\rm NML}}\left(x^{t^{*}_{i+1}}_{t^{*}_{i}+1};{\mathcal P}^{*}_{(i)}\right)$,
this inequality is rewritten as follows:\\
\begin{eqnarray}
&\left(\frac{p_{_{\rm NML}}({\bm x};{\bm {\mathcal P}}^{*})}{p({\bm x};\tilde{{\bm {\mathcal P}}})}\right)^{\frac{1}{2}}
\exp \left[\frac{1}{2}\left\{\sum ^{\tilde{m}}_{j=0}\left(
\log C_{\tilde{t}_{j}-\tilde{t}_{j-1}+1} (\tilde{{\mathcal P}}_{(j)})\right)
+\ell (\tilde{{\bm {\mathcal P}}})+n\epsilon \right\}\right]\geq 1 \label{e1n}. \\ \nonumber
\end{eqnarray}

Then by multiplying (\ref{e1n}) into Type II probability,
it is evaluated as:\\
\begin{eqnarray}\label{e2n}
&&\sum _{{\bm x}: \Phi _{t}({\bm x})\leq 0 }p({\bm x}; \tilde{{\bm {\mathcal P}}}) \nonumber \\
&&\leq \sum _{{\bm x}: \Phi _{t}({\bm x})\leq 0 }
\left(p_{_{\rm NML}}({\bm x};{\bm {\mathcal P}}^{*})p({\bm x};\tilde{{\bm {\mathcal P}}})\right)^{\frac{1}{2}}
\exp \left[\frac{1}{2}\left\{\sum ^{\tilde{m}}_{j=0}\left(
\log C_{\tilde{t}_{j}-\tilde{t}_{j-1}+1} (\tilde{{\mathcal P}}_{(j)})\right)
+\ell (\tilde{{\bm {\mathcal P}}})+n\epsilon \right\}\right] \nonumber \\
&&\leq
\exp \left[-nd_{B}^{(n)}({\bm P}^{*}, \tilde{{\bm {\mathcal P}}}) +\frac{1}{2}\left\{\sum ^{\tilde{m}}_{j=0}\left(
\log C_{\tilde{t}_{j}-\tilde{t}_{j-1}+1} (\tilde{{\mathcal P}}_{(j)})\right)
+\ell (\tilde{{\bm {\mathcal P}}})+n\epsilon \right\}\right].\\ \nonumber
\end{eqnarray}
Here we have used the following relation following the definition of Bhattacharyya distance: \\
\begin{eqnarray*}\label{c1}
\sum _{{\bm x}: \Phi _{t}({\bm x})\leq 0 }\left(p({\bm x}; {\bm {\mathcal P}}^{*}_{_{\rm NML}})
p({\bm x}; \tilde{\bm {\mathcal P}})\right)^{\frac{1}{2}}
&\leq &\sum _{{\bm x}}\left(p({\bm x}; {\bm {\mathcal P}}^{*}_{_{\rm NML}})
p({\bm x}; \tilde{\bm {\mathcal P}})\right)^{\frac{1}{2}}\nonumber \\
&=&\exp (-nd_{B}^{n}({\bm {\mathcal P}}_{_{\rm NML}}, \tilde{\bm {\mathcal P}})).\\ \nonumber
\end{eqnarray*}

Under the assumption as in Theorem \ref{epn},
the remaining terms  in (\ref{e2n}) is asymptotically evaluated as follows:
\begin{eqnarray} \nonumber \\
\exp (\tilde{m}{\rm Ddim}(\tilde{\mathcal P}_{(0)}\otimes \cdots \otimes \tilde{\mathcal P}_{(\tilde{m})})(1+o(1))\log n+\ell (\tilde{\bm {\mathcal P}})/2+n\epsilon /2). \label{c2n}
\end{eqnarray}\\
where $\tilde{\mathcal P}_{(0)}\otimes \cdots \otimes \tilde{\mathcal P}_{(\tilde{m})}$ is model concatenation of $\tilde{\mathcal P}_{(0)}, \cdots \tilde{\mathcal P}_{(\tilde{m})}$ with ratio $(\log \tilde{t}_{1}: \cdots :\log (n-\tilde{t}_{m}))$.
Plugging 
(\ref{c2n}) into (\ref{e2n}) yields (\ref{type2n}).
It completes the proof of Theorem \ref{epn}.
\hspace*{\fill}$\Box$\\

By Theorem \ref{epn}, the MDL test has turned out to be effective for determining whether any given $\{m^{*}, {\bm t}^{*},  {\bm {\mathcal P}}^{*}\}$ is a model sequence that generates ${\bm x}$ or not.
This testing is conducted by comparing the total codelength with the minimum encoding length.
Since it holds for any $\{m^{*}, {\bm t}^{*},  {\bm {\mathcal P}}^{*}\}$,
 we can think of the model sequence:\\
\begin{eqnarray}\label{dmsn}
\hat{{\bm {\mathcal P}}}
=\argmin _{{\bm {\mathcal P}}}\{{\mathcal L}({\bm x};
{\bm {\mathcal P}})+\ell ({\bm {\mathcal P}})\},\\  \nonumber
\end{eqnarray}
as the most probable one for generation of ${\bm x}$.
The model sequence selection (\ref{dmsn}) is called {\em dynamic model selection}~(DMS)~\cite{ym05,ym07}.
It also gives a strategy for model change detection on the basis of the MDL principle.
Therefore, Theorem \ref{epn} also gives a justification of DMS.

Yamanishi and Fukushima~\cite{yamanishi19} considered the following more simple hypothesis testing setting:
For $1<t<n$, for ${\mathcal P}_{(0)}, {\mathcal P}_{(1)},{\mathcal P}_{(2)}\in {\mathcal F}$,
\begin{eqnarray*}\\
H_{0}&: & {\bm x}\sim {\mathcal P}_{(0)},\\
H_{1}&:&  {\bm x}_{+}\sim {\mathcal P}_{(1)}, \  {\bm x}_{-}\sim {\mathcal P}_{(2)}\ \ ({\mathcal P}_{(1)}\neq {\mathcal P}_{(2)}),
\end{eqnarray*}\\
where ${\bm x}_{+}=x_{1},\dots ,x_{t}$ and ${\bm x}_{-}=x_{t+1},\dots , x_{n}$.
We do not know ${\mathcal P}_{(0)}, {\mathcal P}_{(1)}$ nor ${\mathcal P}_{(2)}$ in advance but know ${\mathcal F}$ only.
$H_{0}$ is a hypothesis that $t$ is not a change point while $H_{1}$ is the composite hypothesis that $t$ is a change point.
 They proposed the MDL change statics of the following form:
  For $\epsilon >0$, \\
\begin{eqnarray}\label{mdlchangestatistics}
&&\Phi _{t}({\bm x})\buildrel \rm def \over =\min _{{\mathcal P}\in {\mathcal F}}\{
L_{_{\rm NML}}({\bm x}; {\mathcal P})+\ell ({\mathcal P})\}  \\
&&\ \ \ \ \  -\min _{{\mathcal P}', {\mathcal P}''\in {\mathcal F}}
\bigl\{ L_{_{\rm NML}}({\bm x}_{+};{\mathcal P}')
+L_{_{\rm NML}}({\bm x}_{-};{\mathcal P}'')+\ell ({\mathcal P}', {\mathcal P}'')\bigr\} -n\epsilon , \nonumber \\
\end{eqnarray}
where ${\bm x}_{+}=x_{1},\dots ,x_{t}$ and ${\bm x}_{-}=x_{t+1},\dots , x_{n}$.
$L_{_{\rm NML}}({\bm x};{\mathcal P})$ is the NML codelength for ${\bm x}$ as in (\ref{sc0}) and $\epsilon$ is an accuracy parameter.
$\ell ({\mathcal P})(>0)$ and $\ell ({\mathcal P}',{
\mathcal P}'')(>0)$ are the codelength functions satisfying the Kraft's inequality: \\
\[\sum _{{\mathcal P}\in {\mathcal F}}e^{-\ell ({\mathcal P})}\leq 1,\
\sum _{{\mathcal P}',{\mathcal P}''\in {\mathcal F}}e^{-\ell ({\mathcal P}',{\mathcal P}'')}\leq 1. \]
The MDL change statistics in this case is the difference between the NML codelength without model change and that with model change.
Then we also define the MDL test in this case as follows: We accept ${\mathcal H}_{1}$ if $\Phi _{t}({\bm x})>0$ otherwise
accept ${\mathcal H}_{0}$.
Intuitively, only if the data can be compressed significantly more by changing the distribution at time $t$,
then that point may be thought of as a change point.

The difference between the problem setting in \cite{yamanishi19} and that in Theorem \ref{epn} is that the former is concerned with whether a specific single point is a change point or not, while the latter is concerned with a given model sequence with multiple change points is a true distribution or not.
The latter does not simply include the former as a special case.

Yamanishi and Fukushima~\cite{yamanishi19} derived
upper bounds on error probabilities for the MDL for this scenario. We restate it here so that the error probabilities are related to Ddim as follows:
\begin{theorem}\label{ep}{
Under the conditions as in Theorem \ref{basic} or \ref{basic2},
Type I error probability for the MDL test is 
upper-bounded by\\
\begin{eqnarray}\label{type1}
n^{2{\rm Ddim}({\mathcal P}_{0})(1+o(1))}\exp (-n\epsilon),
\end{eqnarray}\\
while 
Type II error probability for the MDL test is 
upper-bounded by \\
\begin{eqnarray}\label{type2}
n^{2{\rm Ddim}({\mathcal P}_{1}\otimes {\mathcal P}_{2})(1+o(1))}
\exp (-n(d_{B}^{n}(p_{_{{\mathcal P}_{1}\otimes {\mathcal P}_{2}}},\bar{p})-\epsilon /2)),
\end{eqnarray}\\
where ${\mathcal P}_{1}\otimes {\mathcal P}_{2}$ is model concatenation of ${\mathcal P}_{1}$ and ${\mathcal P}_{2}$ with ratio $(\log t: \log (n-t))$.
$d_{B}^{(n)}(p_{_{{\mathcal P}_{1}\otimes {\mathcal P}_{2}}},\bar{p})$ is the Bhattacharyya distance
between $\hat{p}$ and $p^{*}$: \\
\begin{eqnarray*}d_{B}^{(n)}(p_{_{{\mathcal P}_{1}\otimes {\mathcal P}_{2}}},\bar{p})\buildrel \rm def \over =-\frac{1}{n}\log \sum _{{\bm x}} (p({\bm x}:{\mathcal P}_{1}\otimes {\mathcal P}_{2})\bar{p}({\bm x}))^{\frac{1}{2}},
\end{eqnarray*}\\
where $p({\bm x}: {\mathcal P}_{1})\otimes p({\bm x}: {\mathcal P}_{2})=p({\bm x}_{+}:{\mathcal P}_{1})p({\bm x}_{-}:{\mathcal P}_{2})$ is the true distribution for the hypothesis ${\mathcal H}_{1}$ and
\begin{eqnarray}\label{b1}
\bar{p}({\bm x}) &\buildrel \rm def \over =&\max _{{\mathcal P}\in {\mathcal F}}\exp (-L_{_{\rm NML}}({\bm x}; {\mathcal P})-\ell ({\mathcal P}))/C_{n}({\mathcal F})\\
C_{n}({\mathcal F}) &\buildrel \rm def \over =&\sum _{{\bm y}}\max _{{\mathcal P}\in {\mathcal F}}\exp (-L_{_{\rm NML}}({\bm y}; {\mathcal P})-\ell ({\mathcal P})). \label{b2}
\\ \nonumber
\end{eqnarray}

}
\end{theorem}

Theorem \ref{ep} implies that Type I error probability for the MDL test is governed by Ddim for the true model ${\mathcal P}_{0}$ for the hypothesis ${\mathcal H}_{0}$ (see (\ref{type1})),
while its Type II error probability is governed by  Ddim for model concatenation ${\mathcal P}_{1}\otimes {\mathcal P}_{2}$ for the composite hypothesis ${\mathcal H}_{1}$~(see (\ref{type2})).

The small differences between Theorem \ref{ep} and the previous result in \cite{yamanishi19} are: I) in Theorem \ref{ep} the difference between $\bar{p}$ and $p_{_{{\mathcal P}_{1}\otimes {\mathcal P}_{2}}}$ is measured in terms of Bhattacharyya distance, while in the previous work it is measured in terms of $\alpha$-divergence.
II) in Theorem \ref{ep} the error exponents are characterized by {\em Ddim for model concatenation}, while in the previous work, they are characterized by the parametric complexities only.  We omit the proof of Theorem \ref{ep} since it can be proven in the same way as in \cite{yamanishi19} and Theorem \ref{epn}.

Through the analysis both for Theorems \ref{epn} and \ref{ep}, we see that the error probabilities of hypothesis testing for model change detection in two different scenarios are intrinsically related to the parametric complexities of the true models, eventually to their Ddim.

\section{Conclusion}
This paper has introduced a novel notion of model dimensionality, which we call the descriptive dimension (Ddim), from an information-theoretic viewpoint.
Ddim coincides with the parametric dimensionality when the model class is a single parametric one.
It has further been extended to the real-valued dimensionality for general model classes. The paper has specifically considered how to calculate Ddim when a number of models are fused or concatenated.
The paper has derived upper bounds on the rate of convergence of the NML distribution with model of the shortest NML codelength.
The resulting bounds have been characterized by the parametric complexity, eventually by Ddim for model fusion.
The paper has also addressed the issue of hypothesis testing of multiple model change detection with the  MDL test. The paper has derived upper bounds on the error probabilities for the MDL test. They have been characterized by Ddim for model concatenation.
Through the analysis, we have demonstrated that Ddim is an intrinsic quantity which characterizes the  performance of  MDL-based learning and change detection.

Applications of Ddim to  real data mining issues have remained for future studies. Specifically, {\em model change sign detection}  can be thought of one of big challenges. We consider the situation where the underlying model for a data stream changes over time.
Model fusion and concatenation may occur in the transition period of the model changes.
Then the transition period of model changes may be understood through Ddim for model fusion and concatenation.
We expect that signs or early signals of model changes may be tracked by looking at Ddim in the model transition period.
Applications of Ddim to model change sign detection will be dealt with in future work.

\appendix

\section{Proofs}
\subsection{Proof of Theorem \ref{basic}}

Let ${\mathcal P}$ be a $k$-dimensional parametric class, which we denote as ${\mathcal P}_k=\{ p({\bm x};\theta ):\  \theta \in \Theta _{k}\}$ where $\Theta _k$ is a $k$-dimensional parametric space. In this case, we denote $g(\hat{\theta}, \theta )$ instead of $g(\hat{p}, p)$.
Let the finite set of  $k$-dimensional real-valued parameters space be
$\overline{\Theta} _{k}=\{\theta _{1},\theta _{2},\dots\}$ and let
$\overline{{\mathcal P}}_{k}=\{ p({\bm x};\theta ):\  \theta \in \overline{\Theta} _{k}\subset {\mathbb R}^{k}\}.$
Let $I_{n}(\theta )$ be the Fisher information matrix at $\theta$: $I_{n}(\theta )\buildrel \rm def \over =(1/n)E_{\theta}[\partial ^{2}(-\log p({\bm x}; \theta ))/\partial \theta \partial \theta ^{\top}]$ and  suppose that $\lim _{n\rightarrow \infty}I_{n}(\theta )=I(\theta )$ for each $\theta$.
Below we denote $p({\bm x};\theta)$ as $p_{\theta}$.
Consider
\begin{eqnarray*}D_{\epsilon}(i)\buildrel \rm def \over =\{\theta : d_{n}(p_{\theta _{i}}, p_{\theta})\leq \epsilon ^{2}\}.
\end{eqnarray*}\\
Note that $d_{n}(p_{\theta _{i}},p_{\theta})$ is written using Taylor's expansion up to the second order
as follows: Under the condition that $\log p$ is three-times differentiable, $\max _{a,b,c}|\partial ^{3}\log p({\bm x};\theta )/\partial \theta _{a}\partial \theta _{b}\partial \theta _{c}|<\infty$, \\
\begin{eqnarray*}
d_{n}(p_{\theta _{i}},p_{\theta})
& =&\frac{1}{n}E_{\theta _{i}}[\log p({\bm x};\theta _{i})]\\
& &-\frac{1}{n}E_{\theta _{i}}[\log p({\bm x};\theta _{i})]-\frac{1}{n}E_{\theta_{i}}\left[\left.\frac{\partial \log p({\bm x};\theta)}{\partial \theta}\right|_{\theta_{i}}\right](\theta -\theta _{i}) \\
& & +\frac{1}{2n}(\theta -\theta _{i}) ^{\top}E_{\theta _{i}}\left[ \left.-\frac{\partial ^{2}\log p({\bm x};\theta )}{\partial \theta \partial \theta ^{\top}}\right|_{\theta_{i}}\right](\theta -\theta _{i}) +O(||\theta-\theta _{i}||^{3})\\
& =&\frac{1}{2}(\theta-\theta  _{i})^{\top}I_{n}(\theta _{i})(\theta -\theta _{i})+O(||\theta-\theta_{i}||^{3}),
\end{eqnarray*}\\
where we have used the fact:
\begin{eqnarray*}
E_{\theta_{i}}\left[\left.\frac{\partial \log p({\bm x};\theta)}{\partial \theta}\right|_{\theta_{i}}\right]=\sum _{\bm x}\left.p({\bm x};\theta _{i})\frac{\partial \log p({\bm x};\theta )}{\partial \theta }\right|_{\theta_{i}}=\left.\frac{\partial \sum _{\bm x}p({\bm x};\theta )}{\partial \theta}\right|_{\theta_{i}}=0
\end{eqnarray*}
Therefore, we may consider $\tilde{D}_{\epsilon}(i)$ in place of $D_{\epsilon}(i)$.\\
\begin{eqnarray*}
\tilde{D}_{\epsilon}(i)=\{\theta : (\theta -\theta _{i})^{\top}I_{n}(\theta _{i})(\theta -\theta _{i})\leq C\epsilon ^{2}\},
\end{eqnarray*}\\
where $C$ does not depend on $n$ nor $\epsilon$.
Let $B_{\epsilon}(i)$ be the largest hyper-rectangle within $\tilde{D}_{\epsilon}(i)$ centered at $\theta _{i}$.
For some $1\leq  C'< \infty$, for any $i$, we have \\
\begin{eqnarray}\label{volume}
|B_{\epsilon}(i)|\leq |\tilde{D}_{\epsilon}(i)|\leq C'|B_{\epsilon}(i)|\\ \nonumber
\end{eqnarray}

Along with  \cite{rissanen} (p.74), geometric analysis of $B_{\epsilon}(i)$ yields the Lebesgue volume of $B_{\epsilon}(i)$ as follows:
\begin{eqnarray*}
|B_{\epsilon}(i)|&=&\left( \frac{4C\epsilon ^{2}}{k}\right)^{\frac{k}{2}}|I_{n}(\theta _{i})|^{-\frac{1}{2}}=2^{k}\prod ^{k}_{j=1}\sqrt{\frac{C\epsilon ^{2}}{k\lambda_ {j}}},
\end{eqnarray*}
where $\lambda _{j}$ is the $j$-th largest eigenvalue of $I_{n}(\theta _{i})$.

We choose $\overline{\Theta }_{k}$ so that the central limit theorem holds
in the form of (\ref{clt}),  for sufficiently large $n$, as $\theta \rightarrow \theta _{i}$,\\
\begin{eqnarray*}
g(\theta _{i}, \theta )\rightarrow \left( \frac{n}{2\pi}\right)^{\frac{k}{2}} |I(\theta _{i})|^{\frac{1}{2}}.
\end{eqnarray*}\\
Thus we obtain\\
\begin{eqnarray}\label{gfunc2}
g(\theta _{i},\theta_{i})|B_{\epsilon}(i)|\rightarrow \left(
\frac
{2C\epsilon ^{2}n}{k\pi}\right)^{\frac{k}{2}}.
\end{eqnarray}

Next define $Q_{\epsilon}(i)$ as\\
\begin{eqnarray*}
Q_{\epsilon}(i)\buildrel \rm def \over =\int _{\hat{\theta}\in \tilde{D}_{\epsilon}(i)}g(\hat{\theta}, \hat{\theta})d\hat{\theta}.
\end{eqnarray*}\\
and let
$m_{n}(\epsilon )$ be the smallest number of elements in $\overline{{\Theta}}_{k}$:\\
\begin{eqnarray}\label{vv}
\log C_{n}(k)\leq
\log \sum ^{m_{n}(\epsilon )}_{i=1}Q_{\epsilon}(i).
\end{eqnarray}\\
Combining (\ref{gfunc2}) and (\ref{volume}) with (\ref{vv}) yields\\
\begin{eqnarray}
\log C_{n}({\mathcal P}_k)&=&\log m_{n}(\epsilon )+\sup _{\overline{\Theta }_{k}}\left\{\frac{k}{2}\log \left( \frac{2C\epsilon ^{2}n}{k\pi}\right)\right\}+O(1) \label{o1} \\
&=&\log m_{n}(\epsilon )+\frac{k}{2}\log ({\epsilon ^{2}n}) +O(1).\label{o2}
\end{eqnarray}\\
where $C$ in (\ref{o1}) depends on $\overline{\Theta}_{k}$  and the $O(1)$ term in (\ref{o2}) may depend on $k$, but both of them do not depend on $n$ nor $\epsilon$.
The supremum in (\ref{o1}) is taken with respect to $\bar{\Theta}_{k}$ so that (\ref{gfunc2}) holds.
This yields (\ref{nmb1}).
This completes the proof of Theorem \ref{basic}.
\hspace*{\fill}$\Box$

\subsection{Proof of Theorem \ref{rate2n}}
First note that\\
 \begin{eqnarray}\label{totalprob}
 Prob[d^{(n)}_{B}(\hat{p},p^{*})>\epsilon]&\leq &
 Prob[d^{(n)}_{B}(\hat{p},p^{*})>\epsilon |A_{n,\epsilon}]+
 Prob[A_{n,\epsilon}^{c}],\\ \nonumber
 \end{eqnarray}
 where $A_{n,\epsilon}^{c}$ is the complementary set of $A_{n,\epsilon}$.

Let $\hat{{\mathcal P}}$ be the model selected by the MDL learning algorithm and let
$p_{_{\rm NML}}({\bm x};\hat{{\mathcal P})}$ be the NML distribution associated with $\hat{{\mathcal P}}$. We write it as $\hat{p}$.

Let $\tilde{p}=\argmin_{p\in {\mathcal P}}D(p^{*}||p)$.
By the definition of the MDL learning algorithm,
we have\\
\begin{eqnarray}
\min_{{\mathcal P}}( -\log p_{_{\rm NML}}({\bm x}; {\mathcal P}))
& \leq &-\log p_{_{\rm NML}}({\bm x}; {\mathcal P})\nonumber \\
& =&-\log \max _{p\in {\mathcal P}}p({\bm x})+\log C_{n}({\mathcal P})\nonumber \\
& \leq &
-\log p^{*}({\bm x})+\log \frac{p^{*}({\bm x})}{\tilde{p}({\bm x})}+\log C_{n}({\mathcal P}) \nonumber\\
& \leq &-\log p^{*}({\bm x})+nD(p^{*}|| \tilde{p})+\log C_{n}({\mathcal P}) +\frac{n\epsilon }{2}. \label{et0} \\ \nonumber
\end{eqnarray}
Let $J_{n}(p^{*})\buildrel \rm def \over =\min _{{\mathcal P}}\left\{nD(p^{*}|| \tilde{p})+\log C_{n}({\mathcal P}) \right\}$.
Since (\ref{et0}) holds for an arbitrary ${\mathcal P}\in {\mathcal F}$, we have\\
\begin{eqnarray}\label{inn}
\min_{{\mathcal P}}( -\log p_{_{\rm NML}}({\bm x}; {\mathcal P}))&\leq &-\log p^{*}({\bm x})+J_{n}(p^{*})+n\epsilon . \\ \nonumber
\end{eqnarray}

Let $p_{_{\rm NML},{\mathcal P}}$ be the NML distribution $p_{_{\rm NML}}({\bm x}:{\mathcal P})$ associated with ${\mathcal P}$ defined as\\
\[p_{_{\rm NML}}({\bm x}:{\mathcal P})=\frac{\max _{p\in {\mathcal P}}p({\bm x})}{C_{n}({\mathcal P})}.\\ \]
For $\epsilon >0$, the following inequalities hold:
\\
\begin{align}
&Prob[d_{B}^{(n)}(\hat{p},p^{*})>\epsilon |A_{n,\epsilon}]Prob[A_{n,\epsilon}] \nonumber \\
&\leq
Prob[ {\bm x}: \ (\ref{inn})\ {\rm holds\ under}\ d_{B}^{(n)}(\hat{p},p^{*})>\epsilon  |A_{n,\epsilon}]Prob[A_{n,\epsilon}]  \nonumber \\
&\leq
Prob[ {\bm x}: \ (\ref{inn})\ {\rm holds\ under}\ d_{B}^{(n)}(\hat{p},p^{*})>\epsilon  ] \nonumber \\
&= Prob\left[ {\bm x}: \min _{{\mathcal P}:d_{B}^{n}(p_{_{\rm NML},{\mathcal P}},p^{*})>\epsilon }(-\log p_{_{\rm NML}}({\bm x};{\mathcal P})) 
\leq
-\log p^{*}({\bm x})+J_{n}(p^{*})+{n\epsilon}
  \right] \nonumber \\
&=Prob\left[ {\bm x}: \max _{{\mathcal P}:d_{B}^{n}(p_{_{\rm NML},{\mathcal P}},p^{*})>\epsilon }p_{_{\rm NML}}({\bm x}:{\mathcal P})\geq p^{*}({\bm x})e^{-J_{n}(p^{*})-n\epsilon } \right]\nonumber \\
&\leq \sum _{{\mathcal P}\in {\mathcal F}, d_{B}^{(n)}(p_{_{\rm NML},{\mathcal P}}, p^{*})>\epsilon}Prob\bigl[ {\bm x}:
p_{_{\rm NML}}({\bm x}:{\mathcal P})\geq p^{*}({\bm x})e^{-J_{n}(p^{*})-n\epsilon}\bigr]. \label{sumevent}\\
\nonumber 
\end{align}

Let $E_{n}({\mathcal P})$ be the event that\\
\begin{eqnarray*}
p_{_{\rm NML}}({\bm x}:{\mathcal P})
\geq p^{*}({\bm x})e^{-J_{n}(p^{*})-n\epsilon }.\\
\end{eqnarray*}

Note that under the event $E_{n}({\mathcal P})$,
we have \\
\[1\leq \left(\frac{p_{_{\rm NML}}({\bm x};{\mathcal P})}{p^{*}({\bm x})}\right)^{\frac{1}{2}} e^ {J_{n}(p^{*})/2+n\epsilon /2}
. \]\\
Then under the condition that $d_{B}^{(n)}(p_{_{\rm NML},{\mathcal P}}, p^{*})>\epsilon$, we have\\
\begin{eqnarray}
Prob[E_{n}({\mathcal P})]&=&\sum _{{\bm x}\cdots E_{n}({\mathcal P})}p^{*}({\bm x})\nonumber \\
&\leq &\sum _{{\bm x}\cdots E_{n}({\mathcal P})}p^{*}({\bm x})\left(\frac{p_{_{\rm NML}}({\bm x}; {\mathcal P})}{p^{*}({\bm x})}\right)^{\frac{1}{2}} e^{J_{n}(p^{*})/2+n\epsilon /2} \nonumber \\
&\leq& \left\{\sum_{{\bm y}} (p_{_{\rm NML}}({\bm y};{\mathcal P})p^{*}({\bm y}))^{\frac{1}{2}} \right\} e^{J_{n}(p^{*})/2+n\epsilon /2} \nonumber \\
&< &\exp\left(-\frac{n}{2}\left(\epsilon -\frac{J_{n}(p^{*})}{n}\right)\right) ,
\label{event1}
\end{eqnarray}\\
where  we have used the fact that under $d_{B}^{(n)}(p_{_{\rm NML},{\mathcal P}},p^{*})>\epsilon$, it holds \\
\[\sum _{{\bm y}} (p_{_{\rm NML}}({\bm y};{\mathcal P})p^{*}({\bm y}))^{\frac{1}{2}}< e^{-n\epsilon }.\]\\
Plugging (\ref{event1}) into (\ref{sumevent}) yields\\
\begin{eqnarray}
Prob[d_{B}^{(n)}(\hat{p},p^{*})>\epsilon |A_{n,\epsilon}]Prob[A_{n,\epsilon}]
&\leq &\sum _{{\mathcal P}\in {\mathcal F}, d_{B}^{(n)}(p_{_{\rm NML},{\mathcal P}}, p^{*})>\epsilon}Prob[E_{n}({\mathcal P})] \nonumber \\
&< &\sum _{{\mathcal P}\in {\mathcal F}, d_{B}^{(n)}(p_{_{\rm NML},{\mathcal P}}, p^{*})>\epsilon}\exp\left(-\frac{n}{2}\left(\epsilon -\frac{J_{n}(p^{*})}{n}\right)\right) \nonumber \\
&\leq &\sum _{{\mathcal P}\in {\mathcal F}}
\exp\left(-\frac{n}{2}\left(\epsilon -\frac{J_{n}(p^{*})}{n}\right)\right) \nonumber
\\
&=&
|{\mathcal F}|\exp\left(-\frac{n}{2}\left(\epsilon -\frac{J_{n}(p^{*})}{n}\right)\right). \label{jaws} \\ \nonumber
\end{eqnarray}

As for  the probability $Prob[A_{n,\epsilon}^{c}]$, if for some $B_{n}$, a function of $n$, \\
for any ${\mathcal P}$,
$(1/n)|\log ( p^{*}({\bm x})/\tilde{p}({\bm x})|\leq B_{n}$, we employ the  Hoeffding's inequality to obtain the following formula:\\
\begin{eqnarray*}
Prob[A_{n, \epsilon}^{c}]&\leq &\sum _{{\mathcal P}\in {\mathcal F}}2\exp \left( -\frac{n\epsilon ^{2}}{2B_{n}^{2}}\right)
\nonumber \\
&=&2|{\mathcal F}|\exp \left( -\frac{n\epsilon ^{2}}{2B_{n}^{2}}\right).\\
\end{eqnarray*}

Let $P_{n,\epsilon}\buildrel \rm def \over =Prob[A_{n, \epsilon}^{c}]$. Plugging (\ref{jaws}) into (\ref{totalprob}) yields:\\
\begin{eqnarray*}
Prob[d^{(n)}
_{B}(\hat{p},p^{*})>\epsilon ]&< &\frac{|{\mathcal F}|}{1-P_{n,\epsilon}}\exp \left( -\frac{n}{2}\left( \epsilon -\frac{J_{n}(p^{*})}{n}\right)\right)+P_{n,\epsilon}.
\\
\end{eqnarray*}
This completes the proof of Theorem \ref{rate2n}.
\hspace*{\fill}$\Box$\\

\end{document}